\documentclass[11pt]{article}

\usepackage[preprint]{acl}         

\usepackage{times}
\usepackage{latexsym}
\usepackage{inconsolata}

\usepackage[utf8]{inputenc}
\usepackage[T1]{fontenc}

\usepackage{amsmath}
\usepackage{amssymb}
\usepackage{bm}
\allowdisplaybreaks

\usepackage{booktabs}
\usepackage{multirow}
\usepackage{array}
\usepackage{makecell}
\usepackage{pifont}                
\usepackage{rotating}              
\usepackage{placeins}             
\usepackage{float}                
\usepackage{dblfloatfix}          

\usepackage{longtable}            
\usepackage{multicol}
\raggedcolumns
\usepackage{enumitem}
\usepackage{graphicx}
\usepackage{algorithm}
\usepackage{algpseudocode}

\usepackage{xcolor}
\usepackage{tcolorbox}            
\tcbuselibrary{breakable}         
\definecolor{deep}{HTML}{3F2A6D}

\definecolor{tlbad}{HTML}{991B1B}
\definecolor{tlmid}{HTML}{B26B00}
\definecolor{tlgood}{HTML}{0B7A0B}
\DeclareRobustCommand{\hnone}[1]{\tikz[baseline=-0.45ex]{%
  \draw[#1,line width=0.45pt] (0,0) circle (0.42ex);}}
\DeclareRobustCommand{\hhalf}[1]{\tikz[baseline=-0.45ex]{%
  \fill[#1] (0,-0.42ex) arc (-90:90:0.42ex) -- cycle;
  \draw[#1,line width=0.45pt] (0,0) circle (0.42ex);}}
\DeclareRobustCommand{\hfull}[1]{\tikz[baseline=-0.45ex]{%
  \fill[#1] (0,0) circle (0.42ex);}}
\DeclareRobustCommand{\cbad}[1]{\hnone{tlbad}\,\textcolor{tlbad}{#1}}
\DeclareRobustCommand{\cmid}[1]{\hhalf{tlmid}\,\textcolor{tlmid}{#1}}
\DeclareRobustCommand{\cgood}[1]{\hfull{tlgood}\,\textcolor{tlgood}{#1}}
\definecolor{accup}{RGB}{0,140,0}            
\definecolor{accdn}{RGB}{200,0,0}            
\newcommand{\up}[1]{\textcolor{accup}{#1}}   
\newcommand{\dn}[1]{\textcolor{accdn}{#1}}   

\usepackage{pgfplots}
\pgfplotsset{compat=1.18}
\usepgfplotslibrary{groupplots}
\usetikzlibrary{arrows.meta,positioning,fit,backgrounds,calc}

\definecolor{vizblue}{HTML}{2A78D6}    
\definecolor{vizorange}{HTML}{EB6834}  
\definecolor{vizaqua}{HTML}{1BAF7A}    
\definecolor{vizred}{HTML}{E34948}     
\definecolor{vizgrid}{HTML}{E1E0D9}    
\definecolor{vizaxis}{HTML}{C3C2B7}    
\definecolor{vizmuted}{HTML}{898781}   

\usepackage{microtype}
\usepackage{subfiles}              

\makeatletter
\renewcommand\paragraph{\@startsection{paragraph}{4}{\z@}%
  {0.9ex plus 0.15ex minus 0.1ex}{-1em}{\normalsize\bfseries}}
\makeatother

\usepackage{hyperref}
\hypersetup{
  colorlinks = true,
  linkcolor  = deep,
  citecolor  = deep,
  urlcolor   = deep,
}

\newcommand{\ctx}{\operatorname{ctx}}
\newcommand{\cossim}{\operatorname{cos}}
\newcommand{\relu}{\operatorname{ReLU}}
\newcommand{\softplus}{\operatorname{softplus}}
\newcommand{\softmax}{\operatorname{softmax}}
\newcommand{\Ptrie}{P_{\mathrm{trie}}}
\newcommand{\normR}{\widehat{R}}
\newcommand{\reg}{\mathcal{R}}
\newcommand{\children}{\mathcal{C}}
\newcommand{\api}[1]{\texttt{\small #1}}

\newcommand{\trace}{\mbox{\textsc{Tracer}-fixed}}
\newcommand{\tracer}{\textsc{Tracer}}
\newcommand{\tracers}{\mbox{\textsc{Tracer}+G}}
\newcommand{\llmonly}{\mbox{\textsc{LLM-only}}}   
\newcommand{\vecsearch}{\textsc{Vector}}

\newcommand{\best}[1]{\textbf{#1}}           
\newcommand{\runnerup}[1]{\underline{#1}}    
\newcommand{\mi}{\textsuperscript{$\ddagger$}} 
\newcommand{\ciccell}[2]{\makecell{#2\%\\{\tiny(#1)}}} 
\newcommand{\oodshare}[2]{\makecell{#1\%\\{\tiny(#2)}}}  
\newcommand{\ciccellb}[2]{\makecell{\textbf{#2\%}\\{\tiny(#1)}}} 
\newcommand{\cmark}{\textcolor{accup}{\ding{51}}}   
\newcommand{\xmark}{\textcolor{accdn}{\ding{55}}}   

    \title{\textsc{Tracer}: Trajectory-Aware Tool Retrieval for LLM Agents
      }

\author{%
  Tanish Kher\thanks{Work done during an internship at Microsoft.
  Correspondence to: \texttt{tkher@berkeley.edu}.}
  \quad Sarvesh DK
  \quad Abedelkadir Asi \\
  \bfseries Sparsha Chennarayapatna Sundaresh
  \quad Dezhi Hong \\[0.35em]
  \normalfont Microsoft
}
\date{}

\begin{document}
\maketitle

\begin{abstract}
Modern agents select tools from registries, but exposing entire registries incurs high token costs and can impair selection as they grow. We present \tracer{}, a retrieval front-end that indexes observed tool-call sequences in a trie and ranks continuations using a learned blend of empirical and contextual priors. 
On a mixed retail-and-telecom $\tau^2$ environment,
\tracer{} improves GPT-5.4's tool-selection accuracy by $4.7$ points while
reducing input tokens per step by $80\%$, exposing just $5$ of $128$ tools. In autonomous full-turn execution, it matches GPT-5.4's
full-registry task success with $61\%$ fewer input tokens per step. Across six
environments and four models, \tracer{} improves tool-selection
accuracy over full-registry selection in $21$ of $24$ model-environment pairs, with average gains of $6$ to $10$ points; the three exceptions trail by at most $2.2$. We analyze failures in both regimes to identify directions for future work.
\end{abstract}



\section{Introduction}
\label{sec:intro}

Modern language-model agents use tool calling to resolve user requests
\citep{schick2023toolformer,yao2023react}, often executing multi-turn
sequences of function calls over registries of internal services and
external APIs \citep{barres2025tau2bench}. Systems such as Claude Cowork
\citep{anthropic2026cowork}, Copilot Cowork
\citep{microsoft2026cowork}, and Codex \citep{openai2025codex} illustrate
the growing use of tool-mediated interaction in agentic workflows
\citep{xu2026evolution}. A central challenge is selecting and invoking
the right tool from a potentially large registry. Exposing many tool
schemas simultaneously increases context and token costs, while long,
detailed inputs can make relevant information harder to use reliably
\citep{sakizli2026toolschema,liu2024lost}. To manage this overhead,
recent approaches retrieve or load tools on demand
\citep{fei2025mcpzero,anthropic2026toolsearch}, often using tool
descriptions to identify candidates. However, description-based retrieval
can confuse semantically similar tools, and retrieval errors can propagate
to downstream execution. As registries grow, tool selection and schema
management therefore remain important sources of inefficiency and
unreliability despite progress in tool retrieval and tool-use training
\citep{liu2026toolscope,gillespie2026scaling,qin2024toolllm,patil2023gorilla,gan2025ragmcp}.


\begin{figure}[!t]
  \centering
  \begin{tikzpicture}
    \begin{axis}[
      ybar,
      width=\linewidth, height=4.6cm,
      bar width=20pt,
      enlarge x limits=0.28,
      symbolic x coords={Vector,LLM-only,Tracer},
      xtick={Vector,LLM-only,Tracer},
      xticklabels={{\texttt{Vector Search}\\\texttt{+\,GPT-5.4}},{\texttt{GPT-5.4}\\\texttt{Only}},{\textsc{Tracer}\\\texttt{+\,GPT-5.4}}},
      xticklabel style={align=center, font=\scriptsize, color=vizmuted},
      ymin=0, ymax=74, ytick={0,20,40,60},
      ylabel={Selection accuracy (\%)},
      ylabel style={font=\scriptsize, inner sep=2pt},
      yticklabel style={font=\scriptsize, color=vizmuted},
      axis line style={vizaxis, line width=0.5pt},
      tick style={vizaxis},
      ymajorgrids=true,
      grid style={vizgrid, line width=0.4pt},
      nodes near coords,
      nodes near coords style={font=\scriptsize, color=vizmuted,
        /pgf/number format/.cd, fixed, fixed zerofill, precision=1},
    ]
      \addplot[fill=vizblue, draw=none, bar shift=0pt]
        coordinates {(Vector,27.7)};
      \addplot[fill=vizblue, draw=none, bar shift=0pt]
        coordinates {(LLM-only,57.2)};
      \addplot[fill=vizred, draw=none, bar shift=0pt]
        coordinates {(Tracer,61.9)};
    \end{axis}
  \end{tikzpicture}
  \caption{\textbf{Selection accuracy at a fixed budget of retrieving top-$k(=5)$ tools} (mixed $\tau^2$ Large, $128$ tools, GPT-5.4).}
  \label{fig:barsel}
\end{figure}
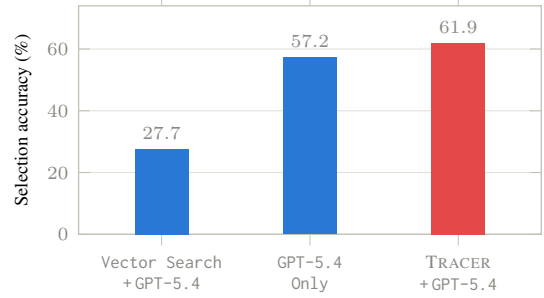

The core gap is that each step is treated as independent search, when tool
retrieval is really sequential prediction: which call comes next, given those
already made. Within a single turn, a
$\tau^2$ retail request is resolved in a logical order: identify the customer,
pull their account, open the order in question, then act on it. The calls
already made therefore constrain the next one.

Built on this intuition, we propose \tracer{}
(\emph{Trie Retrieval Across Contextual Edges, Reweighted}),
a state-conditioned front-end that offers a small, accurate candidate set at
every step and still leaves the final choice to the model
(Figure~\ref{fig:arch}, Appendix~\ref{app:method}). It rests on three ideas. \emph{Full-prefix
conditioning}: execution history is a trie over observed tool-call sequences,
the agent's prefix within a turn is a node, and the tools seen after it are its
candidates, so earlier calls separate next steps that the last call alone leaves
equally plausible. \emph{Transition-specific usage memory}: each edge stores the
conversations that transition was taken in, telling apart candidates whose
descriptions do not. \emph{State-dependent weighting}: history and semantics are
not equally informative, so a learned per-step blend sets how much
each should count at any step. Where a registry's calls concentrate in generic utility tools that no
single state predicts, \tracers{} unions a small global set into every candidate
set.

On a mixed $\tau^2$ environment spanning retail and telecom domains, \tracer{}
paired with GPT-5.4 outperforms the standalone frontier GPT-5.4 model by $4.7$
points in tool-selection accuracy, $57.2\%$ to $61.9\%$, all while cutting
prompt tokens per step by $80\%$ (Figure~\ref{fig:barsel}). Running a full turn
on its own selections, \tracer{} matches full-registry task success, $58.3\%$ against $58.2\%$,
despite presenting only $23\%$ of the tools at any time and cutting per-step
tokens by $61\%$. Similar results largely hold across five other
environments, four models, and budgets $k$ from $3$ to $30$
(\S\ref{sec:results}).

\paragraph{Contributions.}
\begin{itemize}[leftmargin=1.4em,itemsep=2pt,topsep=3pt]
\item \textbf{A comprehensive state-conditioned retrieval front-end.} We combine
  full-prefix conditioning, transition-specific usage memory and a learned
  per-state blend in a single index.
\item \textbf{A benchmark environment built for retrieval.} We pool retail with
  telecom and add synthetic APIs and trajectories to reach $128$ deliberately
  confusable tools with per-step ground truth; a blind human study judged ours
  more human-like than the $\tau^2$ originals.
\item \textbf{Evidence across six environments, four models, and both regimes.}
  \tracer{} improves selection accuracy over full-registry selection in $21$ of $24$
  model-environment pairs while cutting
  per-step context by $55$--$90\%$, and matches task success at $61\%$ fewer
  prompt tokens per step, the dominant term in serving cost.
\item \textbf{Demonstrated gains in specificity and call ordering.} \tracer{}
  sharpens the model's use of specific tools by surfacing similar past
  trajectories, removing competitor tools it would otherwise have picked and
  improving adherence to required procedures.
\end{itemize}

\section{Related Work}
\label{sec:related}

\paragraph{Matching requests to fixed tool representations.}
Retrievers over large registries rank tools by similarity to the request
\citep{qin2024toolllm,patil2023gorilla,gan2025ragmcp,liu2026toolscope,repantis2026howmany}.
Description matchers differ only in how that match is trained
\citep{zheng2024toolrerank,molfetta2025ports,sengupta2025tooldreamer}; Tool2Vec
\citep{moon2024tool2vec} matches past requests rather than documentation; and
MCP-Zero \citep{fei2025mcpzero} has the model write its own query, better
aligned to tool descriptions than the user's words. Each improves the query;
all still score it against a representation fixed in advance. Fixed once, it
cannot separate several plausible candidates, and it cannot register that one
tool serves different purposes at different points. Past usage at that point
does both.

\paragraph{Relating tools to one another.}
TGR \citep{gao2025tgr} smooths each tool's embedding toward the tools it
depends on, so prerequisites are retrieved alongside what needs them; ProTIP
\citep{anantha2023protip} retrieves progressively, subtracting the embeddings
of tools already chosen so the query carries what remains. While these take
real steps, they ignore what makes tool-calling dynamic: every choice must be
read against the calls and turns that came before it. TGR pools dependency
over all prefixes, fixing once a relationship that holds only sometimes, and
ProTIP's subtraction keeps no record of the order the previous calls came in.

\begin{table*}[t]
  \centering
  \caption{\textbf{Positioning \tracer{} against prior tool-retrieval methods.}
    Marks run \hnone{tlbad}~absent, \hhalf{tlmid}~partial,
    \hfull{tlgood}~fullest form.}
  \label{tab:related}
  \scriptsize
  \setlength{\tabcolsep}{3.5pt}
  \renewcommand{\arraystretch}{1.05}
  \begin{tabular}{@{}l c
      >{\raggedright\arraybackslash}p{2.7cm}
      >{\raggedright\arraybackslash}p{2.0cm}
      >{\raggedright\arraybackslash}p{1.95cm} c@{}}
    \toprule
    & \multicolumn{2}{c}{\emph{Evidence from past runs}} & & & \\
    \cmidrule(lr){2-3}
    Method & \makecell{transition\\counts}
           & \makecell{situation match\\{\scriptsize\itshape compared against}}
           & \makecell{Retrieval state\\{\scriptsize\itshape index key}}
           & \makecell{Signals used\\{\scriptsize\itshape count, weighting}}
           & \makecell{LLM\\decides} \\
    \midrule
    ToolRerank \citep{zheng2024toolrerank}
      & \cbad{} & \cbad{tool docs} & \cbad{none}
      & \cbad{1} & \cgood{} \\
    PORTS \citep{molfetta2025ports}
      & \cbad{} & \cbad{tool docs} & \cbad{none}
      & \cbad{1} & \cgood{} \\
    ToolDreamer \citep{sengupta2025tooldreamer}
      & \cbad{} & \cbad{generated docs} & \cbad{none}
      & \cbad{1} & \cgood{} \\
    Tool2Vec \citep{moon2024tool2vec}
      & \cbad{} & \cmid{one user query, per tool} & \cbad{none}
      & \cbad{1} & \cgood{} \\
    MCP-Zero \citep{fei2025mcpzero}
      & \cbad{} & \cbad{server $+$ tool docs} & \cbad{none}
      & \cmid{2, fixed} & \cgood{} \\
    TGR \citep{gao2025tgr}
      & \cbad{} & \cbad{tool docs $+$ deps} & \cbad{none}
      & \cbad{1} & \cgood{} \\
    ProTIP \citep{anantha2023protip}
      & \cbad{} & \cbad{tool docs} & \cmid{implicit plan}
      & \cbad{1} & \cgood{} \\
    \midrule
    ToolNet \citep{liu2024toolnet}
      & \cgood{} & \cbad{tool docs, first step}
      & \cmid{last tool}
      & \cbad{1 per step} & \cgood{} \\
    AutoTool \citep{jia2026autotool}
      & \cgood{} & \cbad{tool docs (SimCSE)} & \cgood{last 2 tools}
      & \cmid{2, fixed $\alpha$} & \cmid{partial} \\
    \midrule
    \textbf{\tracer{}}
      & \cgood{} & \cgood{whole conversation so far, per transition}
      & \cgood{full turn prefix}
      & \cgood{2, learned per state} & \cgood{} \\
    \bottomrule
  \end{tabular}
\end{table*}

\paragraph{Conditioning on execution.}
ToolNet \citep{liu2024toolnet} offers the successors of the last call,
weighted by bigram counts; AutoTool \citep{jia2026autotool} widens that to
two and blends the count with SimCSE similarity at a constant
$\alpha{=}0.5$, using the result to bypass the model. Neither asks how the
current conversation resembles the ones
that previously took each transition: ToolNet has no such signal, and
AutoTool's compares against a tool's description ($21$ points worse on API-Bank
in our ablation). Both also fix the blend once. \tracer{} improves this by
learning the balance of factors per state instead, worth $0.9$ to $7.6$ points
(Table~\ref{tab:ablation-b}; Table~\ref{tab:related} compares all methods).


\section{Method}\label{sec:method}

\subsection{Problem definition and notation}\label{sec:notation}

An agent answers a request $q$ by emitting an ordered sequence of tool calls; at
step $i$ it has emitted $S_i=(t_1,\dots,t_i)$ and must choose the next call from
a registry $\reg$. We
ask: \emph{given the calls already issued in the current turn, how can we use
historical successful trajectories to select $k\ll|\reg|$ tools that contain the
gold tool the agent should call next, cutting per-step token cost while improving
selection accuracy?}

\subsection{Overview}\label{sec:overview}

We model this with a trie built from a dataset $\mathcal{D}$ of training
trajectories (Figure~\ref{fig:arch}, Appendix~\ref{app:method}). At each step
the front-end reads the conversation context $\ctx(q,S_i)$ and the current
prefix $S_i$, locates $S_i$ in the trie (\S\ref{sec:trie}), and scores that
node's children on two signals: an \emph{empirical prior} over the transitions
observed out of $S_i$ (\S\ref{sec:prior}), and a \emph{contextual prior}
measuring how closely the current conversation resembles the ones in which each
transition was historically taken (\S\ref{sec:likelihood}). A learned per-step
blend (\S\ref{sec:wt}) combines them into one score, and the top $k$ form the
candidate set $\mathcal{K}\subset\reg$ from which the model makes its choice.
Every system we compare changes which tools reach the prompt, but the model
always chooses.

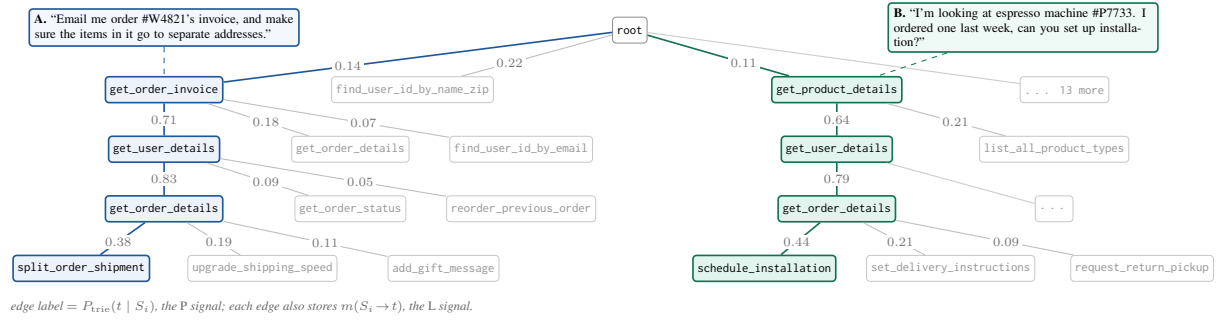
\begin{figure*}[t]
  \centering
  \resizebox{\linewidth}{!}{%
  \begin{tikzpicture}[
    font=\small,
    tn/.style={rounded corners=2pt, draw=black!45, fill=white, line width=0.5pt,
               font=\ttfamily\scriptsize, inner sep=2.5pt,
               minimum height=5mm},
    tnA/.style={tn, draw=vizblue!72!black, fill=vizblue!9,  line width=0.9pt},
    tnB/.style={tn, draw=vizaqua!65!black, fill=vizaqua!13, line width=0.9pt},
    tnfade/.style={tn, draw=black!22, text=black!42},
    ted/.style={draw=black!45, line width=0.6pt},
    edA/.style={draw=vizblue!72!black, line width=0.9pt},
    edB/.style={draw=vizaqua!65!black, line width=0.9pt},
    elbl/.style={font=\scriptsize, text=black!55, inner sep=1.5pt, fill=white},
    lbl/.style={font=\scriptsize\itshape, text=black!60},
    q/.style={rounded corners=2pt, fill=white, line width=0.9pt, align=left,
              inner sep=3pt, text width=50mm, font=\scriptsize},
    qA/.style={q, draw=vizblue!72!black, fill=vizblue!5},
    qB/.style={q, draw=vizaqua!65!black, fill=vizaqua!7},
  ]

  \node[tn] (root) at (11.0,0) {root};
  \node[qA] (qa) at ( 2.0,0.05) {\textbf{A.} ``Email me order \#W4821's
     invoice, and make sure the items in it go to separate addresses.''};
  \node[qB] (qb) at (18.6,0.05) {\textbf{B.} ``I'm looking at espresso
     machine \#P7733. I ordered one last week, can you set up installation?''};

  \node[tnA] (a1) at ( 2.00,-1.15) {get\_order\_invoice};
  \node[tnfade] (f1) at ( 6.80,-1.15) {find\_user\_id\_by\_name\_zip};
  \node[tnB] (a2) at (15.00,-1.15) {get\_product\_details};
  \node[tnfade] (f2) at (19.40,-1.15) {\dots\ 13 more};

  \node[tnA] (b1) at ( 2.00,-2.30) {get\_user\_details};
  \node[tnfade] (f3) at ( 5.60,-2.30) {get\_order\_details};
  \node[tnfade] (f3b) at ( 8.90,-2.30) {find\_user\_id\_by\_email};
  \node[tnB] (b2) at (15.00,-2.30) {get\_user\_details};
  \node[tnfade] (f4) at (19.20,-2.30) {list\_all\_product\_types};

  \node[tnA] (c1) at ( 2.00,-3.45) {get\_order\_details};
  \node[tnfade] (f5) at ( 5.60,-3.45) {get\_order\_status};
  \node[tnfade] (f5b) at ( 8.90,-3.45) {reorder\_previous\_order};
  \node[tnB] (c2) at (15.00,-3.45) {get\_order\_details};
  \node[tnfade] (f6) at (19.20,-3.45) {\dots};

  \node[tnA] (d1) at ( 0.40,-4.60) {split\_order\_shipment};
  \node[tnfade] (d2) at ( 3.90,-4.60) {upgrade\_shipping\_speed};
  \node[tnfade] (d2b) at ( 7.40,-4.60) {add\_gift\_message};
  \node[tnB] (d3) at (13.60,-4.60) {schedule\_installation};
  \node[tnfade] (d4) at (17.20,-4.60) {set\_delivery\_instructions};
  \node[tnfade] (d4b) at (20.90,-4.60) {request\_return\_pickup};

  \draw[edA] (root) -- node[elbl,pos=0.68]{$0.14$} (a1);
  \draw[ted,black!25] (root) -- node[elbl,pos=0.68]{$0.22$} (f1);
  \draw[edB] (root) -- node[elbl,pos=0.68]{$0.11$} (a2);
  \draw[ted,black!20] (root) -- (f2);
  \draw[edA] (a1) -- node[elbl,pos=0.5]{$0.71$} (b1);
  \draw[ted,black!25] (a1) -- node[elbl,pos=0.58]{$0.18$} (f3);
  \draw[ted,black!25] (a1) -- node[elbl,pos=0.62]{$0.07$} (f3b);
  \draw[edB] (a2) -- node[elbl,pos=0.5]{$0.64$} (b2);
  \draw[ted,black!25] (a2) -- node[elbl,pos=0.58]{$0.21$} (f4);
  \draw[edA] (b1) -- node[elbl,pos=0.5]{$0.83$} (c1);
  \draw[ted,black!25] (b1) -- node[elbl,pos=0.58]{$0.09$} (f5);
  \draw[ted,black!25] (b1) -- node[elbl,pos=0.62]{$0.05$} (f5b);
  \draw[edB] (b2) -- node[elbl,pos=0.5]{$0.79$} (c2);
  \draw[ted,black!20] (b2) -- (f6);
  \draw[edA] (c1) -- node[elbl,pos=0.6]{$0.38$} (d1);
  \draw[ted,black!25] (c1) -- node[elbl,pos=0.6]{$0.19$} (d2);
  \draw[ted,black!25] (c1) -- node[elbl,pos=0.62]{$0.11$} (d2b);
  \draw[edB] (c2) -- node[elbl,pos=0.6]{$0.44$} (d3);
  \draw[ted,black!25] (c2) -- node[elbl,pos=0.6]{$0.21$} (d4);
  \draw[ted,black!25] (c2) -- node[elbl,pos=0.62]{$0.09$} (d4b);
  \draw[edA,dashed,line width=0.5pt] (qa) -- (a1);
  \draw[edB,dashed,line width=0.5pt] (qb) -- (a2);

  \node[lbl,anchor=west] at (-1.1,-5.35)
    {edge label $=\Ptrie(t\mid S_i)$, the \emph{P} signal; each edge also
     stores $m(S_i\!\to\!t)$, the \emph{L} signal.};
  \end{tikzpicture}}
  \caption{\textbf{\tracer{}'s execution trie (schematic example).}}
  \label{fig:trie}
\end{figure*}

\subsection{Trie construction}\label{sec:trie}

We build the execution trie by inserting every prefix of every trajectory in
$\mathcal{D}$ as a node. Figure~\ref{fig:trie} shows a hypothetical example.
For any prefix node $S_i$, its children $\children(S_i)$ are the tools observed
immediately after $S_i$ in training.

Each edge from $S_i$ to a child node $t$ carries two kinds of information: the
conditional probability of a training turn traversing $S_i\!\to\!t$, the
\emph{empirical prior} (\S\ref{sec:prior}); and the mean-pooled embedding of
the training contexts in which turns traversed $S_i\!\to\!t$, an \emph{edge
memory} $m(S_i\!\to\!t)$ from which the contextual prior is computed
(\S\ref{sec:likelihood}).

The two are scoped differently by design. The trie is turn-local, resetting to
the root at each turn boundary, which keeps prefixes estimable instead of
fragmenting the state space across a whole conversation; we train on turns drawn
from throughout a conversation so the model sees resets at varied positions. The
edge memory is not turn-local: it embeds the full conversation, so cross-turn
information reaches the retriever without entering the trie as structure.

\subsection{The empirical prior $P(t)$}\label{sec:prior}

Formally, the empirical prior is the normalized transition frequency out of
that node, and at inference we write
$P(t)=\log\big(\Ptrie(t\mid S_i)+\epsilon\big)$ for the resulting log-signal,
adding a small smoothing constant $\epsilon$ so that candidates carrying no
observed transition from this state have a finite value.

The distribution of empirical priors starts more evenly across tools and
sharpens with depth, with some states admitting a single observed successor
(Figure~\ref{fig:arch}, Appendix~\ref{app:method}). The prior therefore steers
the model down the paths training took most often and filters out tools that
never followed $S_i$. The fallback (\S\ref{sec:gate}) deals with
out-of-distribution steps, ensuring the system generalizes to unseen
trajectories.

\subsection{The contextual prior $L(t)$}\label{sec:likelihood}

At every step we embed the step context under a shared, frozen embedder $e$: all
previous turns and step reasonings, with tool-call signatures dropped to limit
input length. For every child we then take the cosine similarity between that
embedding and the edge's memory,
\begin{equation}
R(t)\;=\;\cossim\!\big(e(\ctx(q,S_i)),\,m(S_i\!\to\!t)\big),
\label{eq:R}
\end{equation}
then rectify and normalize across the children into a distribution over them
(uniform if every similarity is non-positive):
\begin{equation}
\normR(t)\;=\;\frac{\relu\!\big(R(t)\big)}{\sum_{t'\in\children(S_i)}\relu\!\big(R(t')\big)},
\label{eq:normR}
\end{equation}
where $\relu(x)=\max(x,0)$, and set $L(t)=\log\normR(t)$.

\subsection{Blending them: the learned scoring rule}\label{sec:wt}

\tracer{} scores each child of the current state by a \emph{learned, per-step}
convex blend of the two log-signals, at a \emph{learned} temperature:
\begin{equation}
s(t)\;=\;\frac{1}{T}\Big[(1-w)\,L(t)\;+\;w\,P(t)\Big]
\label{eq:headline}
\end{equation}
for every $t\in\children(S_i)$, where $L(t)$ is the contextual prior of candidate
$t$, $P(t)$ its empirical prior, the blend weight $w\in(0,1)$, and $1/T>0$.
Children are ranked by $s(t)$ and the top $k$ are presented.

Both $w$ and $T$ are produced per step by a learned map
$f_{\bm\theta}(\bm x)=(w,T)$ over a five-dimensional state feature vector
$\bm x\in\mathbb{R}^5$ computed at each step (\S\ref{app:features}). We represent
$f_{\bm\theta}$ as two linear heads over the standardized feature vector
$\bm z$: a logistic head for $w$, keeping the blend convex, and a floored
softplus head for $T$, keeping the temperature positive and $1/T$ finite
(Algorithm~\ref{alg:step}, Appendix~\ref{app:method}).

We fit $\bm\theta=(\bm a,b_w,\bm c,b_T)$ by minimizing the mean negative
log-likelihood of the gold next call under the softmax over $\children(S_i)$,
with Adam \citep{kingma2015adam} and $\ell_2$ regularization, splitting the
trajectories into inner folds so each step is scored against a trie and edge
memory built without its own trajectory (Eq.~\eqref{eq:nll},
\S\ref{app:method}).

\subsection{Confidence gate and out-of-distribution fallback}\label{sec:gate}

The retriever falls back to global vector search in two cases: an unseen state
($\children(S_i)=\varnothing$) and a flat score distribution ($\kappa<\tau$).
Fallback ranks the full registry by the cosine similarity of the embedded step
context $e(\ctx(q,S_i))$ to each tool's description and returns the top $k$,
replacing the trie candidates entirely. We set $\tau=0.20$. This gate gives the trie-based methods well-defined behavior on
states absent from training, behavior the full-registry and stateless baselines
do not model.

\section{Data and Experimental Setup}
\label{sec:setup}

\subsection{Benchmarks}
\label{sec:setup-benchmarks}

We evaluate on four benchmarks, sweeping the primary one at three registry sizes
(Small, Medium, Large) for six environments in total
(Table~\ref{tab:datastats}). Our primary benchmark,
\textbf{Mixed Tau2 R\&T}, is a
mixed retail\,+\,telecom environment built combining real graded trajectories
from $\tau^2$-bench \citep{barres2025tau2bench} with vetted synthetic ones.
\textbf{Tau3 Banking} is the banking domain of $\tau^3$-Knowledge
\citep{shi2026tauknowledge}, mixing more sophisticated banking tools with
compliance requirements and search over roughly $700$ interconnected documents.
\textbf{BFCL~v3} \citep{patil2025bfcl} is smaller but spans many
weakly-related API families (file-system, vehicle, trading, travel), stressing
cross-domain retrieval. \textbf{API-Bank} \citep{li2023apibank} is the smallest
and most homogeneous, serving as an easy reference point.

We state each change where it arises; Table~\ref{tab:adapt} in
Appendix~\ref{app:adapt} summarizes them. All four satisfy a premise the
trie depends on: every tool is
available at every step, so an edge count reflects what the agent chose rather
than what it was offered.

\subsection{Data sourcing}
\label{sec:setup-sampling}

For the $\tau$ benchmarks we draw real trajectories from the public
leaderboards \citep{sierra2026taubench,shi2026tauknowledge}, keep only
simulations that solved the task and only turns issuing an agent tool call, and
collapse each task to one canonical decision point; Mixed Tau2 R\&T adds
synthetic trajectories (\S\ref{sec:setup-synth}). API-Bank and BFCL~v3 are taken
whole; for BFCL~v3 we verify that merging raises no new alternate answers
(\api{C10}, Appendix~\ref{app:curation}). Trajectory provenance is in
Appendix~\ref{app:prov}; the curation gates are in
Appendix~\ref{app:curation}.


\subsection{Synthetic generation and validation for Mixed Tau2 R\&T}
\label{sec:setup-synthval}
\label{sec:setup-synth}
\label{sec:setup-curation}
\label{sec:setup-human}

Real trajectories alone under-populate the registry, leaving too few plausible
alternatives and making retrieval look artificially easy. We expand each
domain's registry with synthetic APIs and trajectories drawn from a
co-occurrence graph, so that each is a realistic in-context path. Every one
clears blocking gates for ground-truth validity,
non-substitutability against real tools, API disjointness, and non-cloning
($99.8\%$ of retail and $99.9\%$ of telecom clear the last), with accept-sets
seeded where a different \emph{real} tool may legitimately substitute for the
gold call (Table~\ref{tab:curation}, Appendix~\ref{app:curation}).

The diagnostic \api{C3} judge compared real $\tau^2$ trajectories against our
synthetic ones and picked ours as the human-authored one in $57.0\%$ of retail
and $68.9\%$ of telecom pairs. A blind human study agreed: told that exactly one
of each pair was human-authored, three reviewers picked ours in $40$ of $59$
decided pairs, $67.8\%$ of the time (Figure~\ref{fig:fooled},
Appendix~\ref{app:synthtraj}).

\label{sec:setup-policy}Our $\tau^2$-style policy prompts are validated the
same way, so that no clause contradicts any gold trajectory (\api{C6},
Appendix~\ref{app:prompts}).

\subsection{Execution regimes}
\label{sec:setup-regimes}

We evaluate in two execution regimes, differing only in what is appended to
the context between steps.

Teacher-forcing records each selection but feeds the benchmark's gold call
forward in its place, so a wrong selection at step $i$ cannot corrupt step
$i{+}1$. This is our primary regime because it measures the front-end in
isolation, which is what this paper contributes.

Under \emph{free-running}, the agent's own emitted call and its reasoning are
appended and carried forward, so errors compound: a wrong selection changes the
prefix the retriever conditions on. An inference episode ends when our grader confirms the task is complete, or
after a cap of $20$ steps. \label{sec:setup-env}Our grader scores a turn by the
database hash its gold actions produce, plus the read operations a read-only
turn must contain (Appendix~\ref{app:env}).

\subsection{Algorithms}
\label{sec:setup-baselines}

We test \tracer{} against two baselines that ignore execution history and two
variants of our own. \llmonly{} places the entire registry in context at
every step, as a post-disclosure domain list, upper-bounding candidate coverage
at maximal cost. \vecsearch{} ranks the full registry by the cosine similarity
of the embedded step context to each tool's description and returns the top
$k$, the same mechanism the confidence gate falls back to (\S\ref{sec:gate}).

\trace{} is an ablation: it fixes the blend instead of predicting it, scoring
$s(t)=L(t)+P(t)$, and is identical to \tracer{} in every other respect, so the
gap between them isolates the value of the learned blend. \tracers{} is a
configuration for registries whose calls concentrate in context-independent
\emph{utility} tools that no state prior surfaces; it unions a small, fixed,
per-fold global set $G$ into every candidate set after ranking,
$\mathcal{K}=\textsc{TopK}(s,k)\cup G$. $G$ holds the $|G|=3$ tools with the
highest prefix entropy $H(S\mid t)$, the entropy over the states from which $t$
is the gold next call, so it collects the tools whose calls spread most widely
across the trie (\S\ref{app:method}).


\subsection{Models, metrics, and protocol}
\label{sec:setup-protocol}

\paragraph{Models and metrics.} We evaluate four models spanning capability and
vendor: GPT-5.4, GPT-5.4-mini, Grok-4-20-reasoning, and Mistral-Large-3, with
\api{text-embedding-3-small} as the frozen embedder throughout. Under
teacher-forcing we report tool-selection accuracy and context-injection cost
(CIC, the mean prompt tokens injected per decision step); under free-running,
task success rate with CIC. For the few steps with more than one gold tool we
accept alternates (\api{C10}, \S\ref{sec:setup-synthval}).

\paragraph{Protocol.} We teacher-force on every benchmark under five-fold
cross-validation, sweeping the budget $k\in\{5,10,20,30\}$ for Mixed Tau2 R\&T
and $k\in\{3,5,10\}$ for Tau3~Banking, and free-run on the Mixed Tau2 sizes for
GPT-5.4 and GPT-5.4-mini. Cells averaged over three iterations are marked \mi{} in
\S\ref{sec:results}; others are single runs.

\section{Results and Discussion}
\label{sec:results}


\begin{table*}[!t]
  \centering
  \caption{\textbf{Teacher-Forcing tool-selection accuracy (\%) at $k{=}5$ and $k{=}10$.}}
  \label{tab:main-tf}
  \scriptsize
  \setlength{\tabcolsep}{2.5pt}
  \begin{tabular}{@{}ll cc cc cc cc c cc@{}}
    \toprule
    & & \multicolumn{2}{c}{\shortstack{Mixed Tau2\\Small}} & \multicolumn{2}{c}{\shortstack{Mixed Tau2\\Medium}} & \multicolumn{2}{c}{\shortstack{Mixed Tau2\\Large\mi}} & \multicolumn{2}{c}{\shortstack{Tau3\\Banking\mi}} & \shortstack{API-\\Bank$^{\dagger}$} & \multicolumn{2}{c}{\shortstack{BFCL\\v3}} \\
    \cmidrule(lr){3-4}\cmidrule(lr){5-6}\cmidrule(lr){7-8}\cmidrule(lr){9-10}\cmidrule(lr){11-11}\cmidrule(lr){12-13}
    Model & Alg. & $5$ & $10$ & $5$ & $10$ & $5$ & $10$ & $5$ & $10$ & $5$ & $5$ & $10$ \\
    \midrule
    \multirow{5}{*}{GPT-5.4} & \llmonly{} & \runnerup{63.8} (3.1) & 64.1 (3.3) & 60.2 (3.5) & 60.1 (3.4) & 57.2 (1.2) & 57.3 (1.1) & \best{55.0} (2.4) & \best{54.8} (1.6) & 82.4 (6.8) & 61.6 (2.8) & 61.1 (2.8) \\
     & \tracer{} & \best{69.4} (1.5) & \best{69.3} (1.7) & \best{66.1} (2.9) & \best{63.7} (2.6) & \best{61.9} (1.6) & \best{59.0} (1.7) & 45.7 (4.2) & 46.8 (4.2) & 88.3 (7.5) & \best{68.4} (3.7) & \best{68.1} (3.6) \\
     & \tracers{} & 62.4 (1.5) & 62.9 (2.3) & \runnerup{61.9} (3.3) & 60.4 (3.4) & \runnerup{59.9} (1.7) & \runnerup{58.0} (1.8) & \runnerup{51.6} (3.2) & \runnerup{52.6} (3.4) & \runnerup{90.7} (6.2) & \runnerup{65.4} (3.5) & \runnerup{65.6} (3.8) \\
     & \trace{} & 62.2 (1.3) & \runnerup{66.3} (1.8) & 60.3 (2.4) & \runnerup{61.9} (2.8) & 50.5 (1.1) & 51.0 (2.2) & 43.6 (4.4) & 45.9 (4.4) & \best{91.4} (4.5) & 63.1 (2.3) & 64.2 (3.4) \\
     & \vecsearch{} & 26.5 (2.3) & 38.6 (2.2) & 26.1 (1.1) & 34.0 (1.7) & 27.7 (1.5) & 34.4 (1.5) & 9.5 (3.0) & 12.5 (2.7) & 72.7 (5.1) & 56.4 (2.2) & 59.0 (2.9) \\
    \midrule
    \multirow{5}{*}{GPT-5.4-mini} & \llmonly{} & 57.6 (3.1) & 57.2 (3.2) & 55.8 (3.1) & 55.3 (3.4) & 54.5 (1.7) & \runnerup{54.7} (1.7) & \runnerup{49.3} (3.6) & \runnerup{49.4} (3.8) & 81.1 (10.9) & 62.3 (2.6) & 62.9 (1.7) \\
     & \tracer{} & \best{67.5} (1.4) & \best{66.3} (2.3) & \best{62.6} (2.7) & \best{60.4} (2.6) & \best{58.5} (1.8) & \best{55.6} (1.8) & 44.6 (4.0) & 45.2 (3.7) & \runnerup{98.4} (2.0) & \best{70.2} (2.8) & \best{70.0} (2.9) \\
     & \tracers{} & 59.6 (2.0) & 59.9 (2.9) & \runnerup{58.8} (3.0) & 56.6 (3.0) & \runnerup{56.5} (1.5) & 54.4 (1.9) & \best{53.4} (4.7) & \best{52.4} (4.3) & \best{99.2} (1.6) & \runnerup{66.5} (2.2) & \runnerup{67.1} (1.5) \\
     & \trace{} & \runnerup{60.5} (2.3) & \runnerup{63.4} (1.4) & 56.9 (2.7) & \runnerup{57.4} (3.2) & 48.0 (3.4) & 49.7 (3.3) & 42.8 (4.0) & 44.3 (4.2) & 97.6 (1.9) & 63.7 (2.9) & 66.5 (4.0) \\
     & \vecsearch{} & 25.3 (2.7) & 36.0 (2.6) & 26.4 (1.4) & 32.9 (2.5) & 28.4 (1.8) & 33.5 (1.7) & 9.0 (2.8) & 11.1 (2.6) & 76.2 (4.7) & 56.0 (2.6) & 60.1 (3.7) \\
    \midrule
    \multirow{5}{*}{Grok-4-20-r} & \llmonly{} & 62.6 (3.6) & 62.4 (3.5) & 60.5 (4.1) & 60.6 (4.0) & \best{59.8} (1.5) & \best{60.2} (1.2) & \best{58.6} (2.5) & \best{58.7} (2.6) & 86.4 (3.4) & 60.5 (2.9) & 59.8 (2.5) \\
     & \tracer{} & \best{70.4} (2.0) & \best{70.8} (1.9) & \best{66.2} (3.4) & \best{65.1} (2.8) & \runnerup{59.5} (3.8) & \runnerup{58.3} (3.8) & 42.4 (4.3) & 42.7 (4.1) & \runnerup{98.4} (2.0) & \best{71.8} (3.1) & \best{72.4} (3.9) \\
     & \tracers{} & \runnerup{65.7} (2.4) & \runnerup{66.1} (2.4) & \runnerup{63.9} (3.7) & \runnerup{62.8} (3.0) & 58.3 (4.0) & 57.7 (3.6) & \runnerup{56.5} (4.6) & \runnerup{55.7} (4.2) & \best{98.5} (1.9) & \runnerup{70.4} (4.3) & \runnerup{71.3} (4.5) \\
     & \trace{} & 62.6 (1.5) & 65.5 (1.0) & 59.4 (2.5) & 61.3 (2.9) & 48.4 (2.7) & 52.3 (3.2) & 40.4 (4.1) & 42.1 (3.7) & 97.6 (1.9) & 66.8 (4.1) & 65.3 (4.2) \\
     & \vecsearch{} & 25.5 (1.1) & 35.8 (1.3) & 26.6 (0.9) & 34.3 (1.4) & 27.4 (2.6) & 32.8 (2.3) & 8.9 (1.8) & 10.9 (2.6) & 76.4 (1.7) & 60.3 (3.4) & 59.1 (3.1) \\
    \midrule
    \multirow{5}{*}{Mistral-Large-3} & \llmonly{} & \runnerup{66.1} (2.2) & 65.6 (2.2) & 62.6 (3.0) & 62.8 (2.6) & 61.2 (2.0) & \runnerup{61.2} (2.0) & \runnerup{48.0} (2.3) & \runnerup{47.6} (2.2) & 67.2 (7.5) & 62.5 (3.5) & 62.6 (3.8) \\
     & \tracer{} & \best{71.6} (1.4) & \best{71.5} (1.7) & \best{68.3} (2.5) & \best{65.5} (2.3) & \best{64.5} (2.2) & \best{62.2} (2.4) & 43.4 (3.9) & 44.3 (3.9) & \runnerup{95.3} (2.9) & \best{70.1} (3.9) & \best{68.6} (3.4) \\
     & \tracers{} & 65.5 (1.9) & 67.0 (2.0) & \runnerup{64.4} (3.3) & \runnerup{63.5} (3.3) & \runnerup{62.3} (2.1) & 60.5 (2.4) & \best{51.0} (4.0) & \best{51.6} (4.3) & \best{96.1} (3.5) & \runnerup{67.5} (4.8) & \runnerup{68.3} (5.0) \\
     & \trace{} & 64.2 (1.4) & \runnerup{67.2} (1.7) & 60.2 (2.6) & 61.9 (2.8) & 53.6 (1.1) & 54.6 (2.0) & 41.6 (4.6) & 43.2 (4.1) & 93.8 (3.8) & 64.1 (2.4) & 62.8 (3.5) \\
     & \vecsearch{} & 26.5 (2.3) & 39.2 (2.4) & 26.6 (1.3) & 34.1 (1.7) & 27.3 (1.7) & 32.5 (1.7) & 8.7 (2.3) & 10.2 (2.7) & 69.6 (10.1) & 56.9 (1.9) & 57.3 (2.9) \\
    \bottomrule
  \end{tabular}
  \par\smallskip
  {\footnotesize Best per column within each model in \textbf{bold}, second best
  \underline{underlined}. Parentheses give the across-fold SD over the
  five evaluation folds. Full budget sweep in
  Tables~\ref{tab:main-tf-full}--\ref{tab:main-tf-full-b}.}
\end{table*}

Our headline claim is that state-conditioned retrieval matches or exceeds
the full-registry baseline (\llmonly{}) while presenting a small fraction of
the registry.


\subsection{Selection accuracy (Teacher-Forcing)}
\label{sec:results-tf}

Table~\ref{tab:main-tf} reports selection accuracy at $k{=}5$ and $k{=}10$;
the full sweep is Tables~\ref{tab:main-tf-full}--\ref{tab:main-tf-full-b} in
Appendix~\ref{app:fullgrid}.

\tracer{} improves tool-selection accuracy while showing the model a small
subset of the registry. On our primary testbed, Mixed Tau2 R\&T (Large), it
beats the full registry for three of the four models while selecting $5$ of
$128$ tools, by $4.7$ points for GPT-5.4; Grok-4 trails by $0.3$, but this
figure is well inside that cell's $3.8$-point fold SD. The margin generally
widens for every model as the registry shrinks. \tracer{}'s gains exceed one SD in $31$ of
the $34$ cells where it wins; the clearest exception is GPT-5.4 on API-Bank, whose $131$ steps
make the smallest folds. \tracer{}'s stability is also similar to \llmonly{}'s:
their fold SDs differ by a median of $0.8$ points and \tracer{} is tighter in
$21$ of $44$ cells.

Tau3~Banking is where the full registry beats every trie variant for GPT-5.4
and Grok-4, with \tracers{} closest at $3.4$ and $2.1$ points, though
\tracers{} recovers the benchmark outright for the other two models.


\subsection{Context-injection cost}
\label{sec:results-cost}

\begin{figure}[t]
  \centering
  \begin{tikzpicture}
    \begin{axis}[
      width=0.56\linewidth, height=3.8cm, scale only axis,
      xmin=26, xmax=95, ymin=-6.2, ymax=10.6,
      xtick={30,50,70,90},
      xticklabel={\pgfmathprintnumber{\tick}\%},
      ytick={-5,0,5,10},
      xlabel={context saved per decision step (CIC reduction)},
      ylabel={$\Delta$ accuracy (pts)},
      label style={font=\scriptsize}, tick label style={font=\scriptsize},
      ylabel style={inner sep=1pt}, xlabel style={inner sep=1pt},
      tick align=outside, tick pos=left,
      every axis plot/.append style={line width=0.7pt, mark size=1.3pt},
      legend style={at={(0.5,-0.26)}, anchor=north, legend columns=3,
                    draw=vizgrid, fill=white, font=\tiny,
                    inner sep=2pt, row sep=1pt,
                    /tikz/every even column/.append style={column sep=4pt}},
    ]
      \draw[vizaxis, dashed, line width=0.4pt] (axis cs:26,0) -- (axis cs:95,0);
      \node[font=\tiny, text=vizmuted, anchor=west] at (axis cs:26.7,0.9) {parity};
      \node[font=\tiny\itshape, text=vizmuted, anchor=north west]
        at (axis cs:26.7,10.3) {cheaper \emph{and} more accurate};

      \addplot[vizblue, mark=*, -{Stealth[length=6pt, width=5pt]}] coordinates {(55,5.8) (52,5.4)};
      \addlegendentry{Tau2 Small}
      \addplot[vizaqua, mark=*, -{Stealth[length=6pt, width=5pt]}] coordinates {(72,5.4) (69,3.2) (64,2.9)};
      \addlegendentry{Tau2 Medium}
      \addplot[deep, mark=*, -{Stealth[length=6pt, width=5pt]}] coordinates {(80,3.9) (77,1.0) (73,0.5) (68,0.6)};
      \addlegendentry{Tau2 Large}
      \addplot[vizred, mark=*, -{Stealth[length=6pt, width=5pt]}] coordinates {(30,-4.3) (30,-3.4) (29,-2.2)};
      \addlegendentry{Tau3 Banking}
      \addplot[black!55, mark=*, -{Stealth[length=6pt, width=5pt]}] coordinates {(88,9.0) (87,5.9)};
      \addlegendentry{API-Bank}
      \addplot[vizorange, mark=*, -{Stealth[length=6pt, width=5pt]}] coordinates {(90,6.8) (87,7.0) (81,7.1) (76,7.7)};
      \addlegendentry{BFCL v3}
    \end{axis}
  \end{tikzpicture}
  \caption{\textbf{The accuracy--cost trade-off} (GPT-5.4; \tracer{} vs.\
    \llmonly{}, \tracers{} on Tau3~Banking). One path per benchmark, one
    point per budget $k$ up to $25\%$ of the registry; arrows point toward
    larger $k$.}
  \label{fig:tradeoff}
\end{figure}
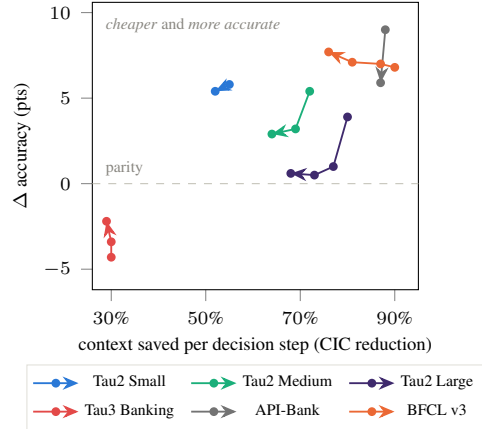

We report \emph{context-injection cost} (CIC), the mean prompt tokens per
decision step, for every benchmark, algorithm and budget in
Tables~\ref{tab:cost-full}--\ref{tab:cost-full-b}, Appendix~\ref{app:fullgrid}.

\tracer{} buys those accuracy gains at a fraction of the context. On our primary
testbed, Mixed Tau2 R\&T (Large), it injects $2.8$k prompt tokens per decision
step against $14.2$k for the full registry, an $80\%$ reduction. As the Mixed
Tau2 pool grows, full-registry cost nearly doubles while \tracer{}'s falls, so
the saving grows from $55\%$ to $80\%$. Across benchmarks the reduction runs
from $31\%$ on Tau3~Banking, whose long conversation histories dominate the
prompt for every front-end, to $90\%$ on BFCL~v3.
Figure~\ref{fig:tradeoff} shows the trade-off directly.

\subsection{End-to-end task success (free-running)}
\label{sec:results-fr}

\begin{table}[t]
  \centering
  \caption{\textbf{Free-running at the best budget per setting.}}
  \label{tab:fr-best}
  \small
  \setlength{\tabcolsep}{2pt}
  \begin{tabular}{@{}ll cccc@{}}
    \toprule
    Registry & Model & \shortstack{Cand.\\set}
      & \shortstack{Margin\\(pts)}
      & \shortstack{Fold\\SD}
      & \shortstack{CIC\\red.} \\
    \midrule
    \multirow{2}{*}{Small} & GPT-5.4          & $43\%$ & $+2.3$ & $4.5$ & $38\%$ \\
     & GPT-5.4-mini     & $22\%$ & $+2.2$ & $5.1$ & $42\%$ \\
    \midrule
    \multirow{2}{*}{Medium} & GPT-5.4          & $33\%$ & $+1.2$ & $2.3$ & $50\%$ \\
     & GPT-5.4-mini     & $33\%$ & $+4.4$ & $3.3$ & $50\%$ \\
    \midrule
    \multirow{2}{*}{Large} & GPT-5.4          & $23\%$ & $+0.1$ & $2.3$ & $61\%$ \\
     & GPT-5.4-mini     & $16\%$ & $+15.2$ & $2.7$ & $68\%$ \\
    \bottomrule
  \end{tabular}
  \par\smallskip
  {\footnotesize Mixed Tau2 R\&T at three registry sizes; full table
   in Table~\ref{tab:main-fr}.}
  \end{table}

The compounding-error nature of free-running makes recall matter more than in
teacher-forcing, so \tracer{} overtakes \llmonly{} at a larger candidate
set (Table~\ref{tab:main-fr}, Appendix~\ref{app:fullgrid}). The share of the
registry it needs, however, falls as the registry grows. On GPT-5.4 it reaches
parity: $+0.1$ points on Large (at $23\%$ of the registry), $+1.2$ on Medium
($33\%$), and $+2.3$ on Small ($43\%$). On GPT-5.4-mini it gains $+13.7$ to
$+15.2$ points on Large across the whole sweep, from $3.9\%$ of the registry
upward.

Those are the best budgets per setting, and the rest of the sweep is less kind.
Of the $24$ model-registry-budget cells we can compare, $16$ sit within $3$
points of \llmonly{} either way (median $2.2$). GPT-5.4 is below \llmonly{} at
$8$ of its $12$ budgets, reaching parity only at the largest.

In these specific situations, selection accuracy gains do not translate
because a single early miss moves the turn onto a prefix the trie has never
seen, and the front-end falls back to vector search, which shows up as a far
higher out-of-trie rate (Table~\ref{tab:ood}, Appendix~\ref{app:ood}). The
tools most exposed are prefix-independent lookups, where \tracer{} already
trails the full registry under teacher forcing since no prefix predicts them.
We return to this in the Conclusion (\S\ref{sec:conclusion}).

For space, and because hyperparameter tuning finds the budget with the
largest task-success margin, we report those gains in
Table~\ref{tab:fr-best}.


\subsection{Why the gains happen}
\label{sec:results-why}

\begin{table}[!t]
  \centering
  \caption{\textbf{The specificity trap} (Mixed Tau2 R\&T Large; GPT-5.4,
    $k{=}5$; strict exact-match).}
  \label{tab:spec-trap}
  \small
  \setlength{\tabcolsep}{3pt}
  \resizebox{\columnwidth}{!}{%
  \begin{tabular}{@{}l r rr@{}}
    \toprule
    Specific API & $N$ & \llmonly{} & \tracer{} \\
    \midrule
    \api{split\_order\_shipment}          & 54 & \dn{14\%} & \up{51\%} \\
    \api{add\_gift\_message}              & 79 & \dn{32\%} & \up{77\%} \\
    \api{request\_return\_pickup}         & 46 & \dn{16\%} & \up{38\%} \\
    \api{schedule\_installation}          & 57 & \dn{29\%} & \up{58\%} \\
    \api{file\_shipping\_insurance\_claim} & 79 & \dn{49\%} & \up{69\%} \\
    \midrule
    Trap rate (lower is better) & 315 & \dn{36.4\%} & \up{16.4\%} \\
    \bottomrule
  \end{tabular}}\\[3pt]
  {\footnotesize \emph{Trap rate} $=$ the share of steps whose gold call is one
   of the five on which the model instead predicted the generic hub
   \api{get\_order\_details}.}
\end{table}

\tracer{} raises selection accuracy for two reasons at once: it surfaces the
gold tool reliably, and the smaller, more relevant set it hands over improves
the executor model's own accuracy on the tools that remain.

\paragraph{Correct Usage of Precision Tools.}
Registries accumulate narrow tools beside a broad one that covers part of the
same request, and picking the narrow one is the whole task. \textbf{Retail}
shows this most sharply. Descriptions alone underdetermine the choice among the
five order-management APIs in Table~\ref{tab:spec-trap}, so \llmonly{} falls
back on the generic hub \api{get\_order\_details}, which on its face covers some
of what is asked, on $36.4\%$ of the steps where one of the five is correct.
\tracer{} cuts the trap rate to $16.4\%$ by separating APIs on the turns they
were called in: at this prefix, the hub was asked for something else.

This filtering matters when the competitor is not merely confusing but
dangerous. For instance, when a customer asked for a prepaid return label,
\llmonly{} incorrectly tried to initiate a return
(\api{return\_delivered\_order\_items}), while \tracer{} ranked that tool $9$th
on the contextual prior at that prefix, dropping
the destructive tool from even being given to the model, so it correctly chose
the precision tool \api{get\_return\_label} (Appendix~\ref{app:cases}).


\begin{figure}[!t]
  \centering
  \renewcommand{\api}[1]{\texttt{#1}}%
  \begin{tikzpicture}[
    font=\scriptsize,
    bub/.style={rounded corners=3pt, inner sep=4pt, align=left, line width=0.4pt},
    usr/.style={bub, fill=vizred!8,    draw=vizred!35},
    pre/.style={bub, fill=black!3,     draw=black!18},
    fe/.style ={bub, fill=black!4,     draw=black!30, dashed},
    cal/.style={bub, fill=vizorange!12,draw=vizorange!55},
    non/.style={bub, fill=black!3,     draw=black!18, font=\scriptsize\itshape},
  ]
    \node[usr, text width=0.86\linewidth, anchor=north] (cust) at (0,0)
      {\textbf{Customer:} ``I'd like to open a business checking account for
       Yumi, and a savings account alongside it.''};
    \node[pre, text width=0.86\linewidth, anchor=north west]
      at ($(cust.south west)+(0,-0.10)$) (pfx)
      {\emph{Steps 0--6 agree.}};

    \coordinate (L) at ($(pfx.south west)+(0,-0.20)$);
    \coordinate (R) at ($(pfx.south west)+(0.455\linewidth,-0.20)$);

    \node[font=\scriptsize\bfseries, anchor=north west] (lh) at (L) {\tracers{}};
    \node[fe,  text width=0.40\linewidth, anchor=north west]
      at ($(lh.south west)+(0,-0.06)$) (lf)
      {\textbf{front-end.} \api{open\_bank\_account}~\xmark\ (no empirical
       prior)\\ \api{unlock:open\_bank\_account}~\cmark};
    \node[cal, text width=0.40\linewidth, anchor=north west]
      at ($(lf.south west)+(0,-0.08)$) (l7)
      {\textbf{step 7.} \api{unlock:open\_bank\_account}~\cmark};
    \node[cal, text width=0.40\linewidth, anchor=north west]
      at ($(l7.south west)+(0,-0.06)$) (l8)
      {\textbf{step 8.} \api{open\_bank\_account}~\cmark\;\;\emph{unlock first,
       then the write}};

    \node[font=\scriptsize\bfseries, anchor=north west] (rh) at (R) {\llmonly{}};
    \node[fe,  text width=0.40\linewidth, anchor=north west]
      at ($(rh.south west)+(0,-0.06)$) (rf)
      {\textbf{front-end.} All $103$ tools placed.};
    \node[cal, text width=0.40\linewidth, anchor=north west]
      at ($(rf.south west)+(0,-0.08)$) (r7)
      {\textbf{step 7.} \api{open\_bank\_account}~\xmark};
    \node[non, text width=0.40\linewidth, anchor=north west]
      at ($(r7.south west)+(0,-0.06)$) (r8)
      {\textbf{step 8.} the account is already open; the unlock is never
       issued};

    \node[fit=(lh)(l8)(rh)(r8), inner sep=0pt] (cols) {};
    \draw[black!30, dashed, line width=0.4pt]
      ($(cols.north)+(0,0.12)$) -- ($(cols.south)+(0,-0.06)$);
  \end{tikzpicture}
  \caption{\textbf{The trie keeps the unlock before the write} (Tau3 Banking,
    GPT-5.4, $k{=}10$).}
  \label{fig:bank-unlock}
\end{figure}
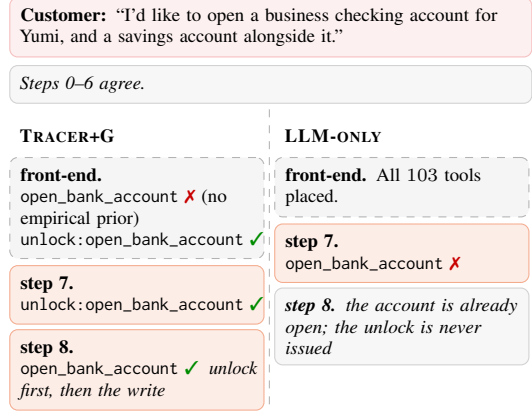

\paragraph{Following Procedural Controls.}
Many deployments gate a consequential action behind a prerequisite call, and
the ordering is itself the requirement. In \textbf{banking}, a sensitive tool must be unlocked by
\api{unlock:}$X$ before it can be called, and under the full registry
\llmonly{} skips that step when asked to open an account. The trie
withholds the raw write: no successful trajectory reaches
\api{open\_bank\_account} from $S_i$ without passing through
\api{unlock:open\_bank\_account}. The edge memory supplies the other half,
matching the conversation to the unlock (Figure~\ref{fig:bank-unlock}).

\subsection{When the global set is needed}\label{sec:globalset}

Adding a global set to \tracer{} (\tracers{}) helps when some tools not bound to
any prefix take a large share of tool calls. This
explains \tracer{}'s low performance on Tau3~Banking, whose registry mostly uses
generic tools that cover many subtasks under one umbrella (\api{KB\_search} for
any knowledge-base lookup, \api{grep} for any text match), so they appear
throughout the trie. \tracers{} recovers $5.8$ points there, landing between
$3.0$ below and $4.0$ above \llmonly{} depending on the model
(Table~\ref{tab:main-tf}). On the $\tau^2$ mixes and BFCL~v3, where usage is
less concentrated or utility tools do not dominate, the injected tools add
distractors rather than recall, and \tracer{} outperforms \tracers{}.

\begin{table}[!t]
  \centering
  \caption{\textbf{The global tool set} (GPT-5.4, $k{=}10$).}
  \label{tab:globalset}
  \scriptsize
  \setlength{\tabcolsep}{3pt}
  \begin{tabular}{@{}l c c c@{}}
    \toprule
    Benchmark & \shortstack{APIs for\\50\% of calls}
      & \shortstack{also in\\top-5 entropy}
      & \shortstack{$\Delta$ accuracy (pts)\\\tracers{} $-$ \tracer{}} \\
    \midrule
    Mixed Tau2 R\&T & 5  & 0 & $-1.0$ \\
    Tau3 Banking    & 2  & 2 & $+5.8$ \\
    BFCL v3         & 17 & 3 & $-2.5$ \\
    \bottomrule
  \end{tabular}
\end{table}


\subsection{Ablations}
\label{sec:ablations}

\paragraph{(A) Edge embeddings vs.\ description-only prior.}
We compare a description-only variant (\api{Trie-Bayesian}) that uses static
tool-description similarity under the same trie and the same fixed blend. In
API-Bank (GPT-5.4-mini), this variant scored $21$ points lower than \trace{}.
The contextual prior moves us from below the \llmonly{} baseline ($81.1$) to
well above it.

\paragraph{(B) Learned vs.\ constant weights (primary).}
Comparing \trace{} against \tracer{} directly tests the paper's core
mechanism: that \emph{state-conditioned} blending beats a globally-tuned
constant. The learned blend helps most in diverse trie maps like
Mixed Tau2 R\&T and BFCL~v3 that lean on the contextual prior over
the empirical prior, and makes little difference in Tau3~Banking where few
generic tools dominate. The head settles on
the contextual prior, averaging $w{=}0.019$ on Mixed Tau2 R\&T (Large),
though its coefficients raise $w$ early in the
conversation and where the empirical prior is already confident
(Tables~\ref{tab:wbyenv}--\ref{tab:wcoef}).

\begin{table}[!t]
  \centering
  \caption{\textbf{Ablation (B): learned vs.\ constant blend} ($k{=}10$).}
  \label{tab:ablation-b}
  \small
  \setlength{\tabcolsep}{4pt}
  \begin{tabular}{@{}l ccc@{}}
    \toprule
    Model & \multicolumn{3}{c}{\shortstack{$\Delta$ accuracy (pts)\\\tracer{} $-$ \trace{}}} \\
    \cmidrule(lr){2-4}
    & \shortstack{Mixed\\Tau2 R\&T} & \shortstack{Tau3\\Banking}
      & \shortstack{BFCL\\v3} \\
    \midrule
    GPT-5.4          & +8.0 & +0.9 & +3.9 \\
    GPT-5.4-mini     & +5.9 & +0.9 & +3.5 \\
    Grok-4-20-r      & +6.0 & +0.6 & +7.1 \\
    Mistral-Large-3  & +7.6 & +1.1 & +5.8 \\
    \bottomrule
  \end{tabular}
\end{table}

\paragraph{(C) Trie key depth.}
We truncated the trie's keys to just the last two calls and ran \tracer{} on
this last-2 trie, which raised selection accuracy by $1.5$ to $2.9$ points on
Grok-4-20-reasoning and Mixed Tau2 R\&T (Large) compared to the full
current-turn prefix (Appendix~\ref{app:last2}). While this is within one SD
($3.7$ to $4.3$), it points to key depth as a parameter worth testing in future
work. The shorter key leaves $9.6\%$ of steps out of trie against $12.4\%$ for
the full prefix, and the two are indistinguishable on the steps that do fall
out. \tracer{} still adds the learned weighting and contextual embeddings that
no bigram front-end has (\S\ref{sec:related}).


\section{Conclusion and Future Work}
\label{sec:conclusion}

\tracer{} is the first tool-retrieval front-end to make selection
history-aware and dynamic at every prefix: it keys on the full execution
prefix, ranks candidates against the conversations each transition was taken
in, and learns how best to weight the two signals for the situation. It beats
full-registry selection in $21$ of $24$ model-environment pairs while cutting
per-step tokens by $80\%$ on our primary testbed, at a cost that tracks the
budget rather than the registry.

The budget $k$ already moves with domain and model
(Table~\ref{tab:fr-best}), and the global set recovers most of the
Tau3~Banking gap; both are fixed here and should adapt.
\tracer{} gained $15.2$ points for GPT-5.4-mini and matched
GPT-5.4's full-registry task success at $23\%$ of the registry, but did not
carry gains at most of GPT-5.4's budgets, largely because running on its own
prefix roughly triples the share of steps that fall outside the trie. Future
work should present the full registry on out-of-trie steps alone, which would
make coverage exact and lift selection accuracy there by $38$ points on
average while retaining $84\%$ of the token reduction
(Table~\ref{tab:ood}), likely translating to higher task success.


\section*{Limitations}
\label{sec:limitations}

\paragraph{Cold start.} \tracer{} indexes a dataset $\mathcal{D}$ of graded
trajectories (\S\ref{sec:trie}), so a registry with no usage history has no
trie to consult and every step falls to the vector fallback, where accuracy is
far below the rest of the run: presenting the full registry on those steps
instead would raise it by $38$ points on average (Table~\ref{tab:ood}). The
method therefore assumes a deployment that has already logged successful runs.
How quickly a trie becomes useful as trajectories accumulate is the obvious
next measurement, and one our fixed-size corpora cannot answer.

\paragraph{Trie key depth.} We key the trie on the full current-turn prefix
(\S\ref{sec:trie}) and did not tune that choice. A shorter key is not obviously
worse (\S\ref{sec:ablations}), so the depth we use may not be the best one.

\paragraph{Free-running episodes are terminated by the grader.} An episode
ends when the grader confirms completion or at a cap of $20$ steps
(\S\ref{sec:setup-env}); the agent never judges for itself that it is done.
Task success therefore measures whether the front-end supplies the tools
needed to reach the goal, not whether an agent built on it would know when to
stop.

\paragraph{Selection accuracy discards ordering freedom.} Several tools can
often be called in different orders and still complete a task, but accuracy is
scored against one recorded ordering, so a valid alternative counts as wrong.
Accept-sets (\S\ref{sec:setup-protocol}) and free-running task success, which
asks only whether the goal was reached, bound the distortion.

\apptocmd{\thebibliography}{%
  \setlength{\itemsep}{1pt plus 1pt minus 0pt}}%
  {\typeout{BIBSEP PATCH OK}}{\typeout{BIBSEP PATCH FAILED}}
\bibliography{references}

\appendix
\onecolumn
\nolinenumbers
\setlength{\intextsep}{12pt plus 4pt minus 4pt}
\makeatletter
\renewcommand\subsection{\@startsection{subsection}{2}{\z@}%
  {-1.1ex plus -0.3ex minus -.2ex}{0.5ex plus .2ex}%
  {\normalsize\bfseries\raggedright}}
\makeatother

\section{Appendix}
\label{sec:appendix}

\subsection{Benchmark statistics}
\label{app:datastats}

\begin{table}[H]
  \centering
  \caption{\textbf{Benchmark statistics.}}
  \label{tab:datastats}
  \small
  \setlength{\tabcolsep}{5pt}
  \resizebox{\textwidth}{!}{%
  \begin{tabular}{@{}l ccc ccc@{}}
    \toprule
    & \multicolumn{3}{c}{Mixed Tau2 R\&T} & Tau3 Banking & BFCL v3 & API-Bank \\
    \cmidrule(lr){2-4}
    & Small & Medium & Large & & & \\
    \midrule
    Domains                          & 2 & 2 & 2 & 1 & 4 & 1 \\
    Tasks (trajectories)             & 311 & 725 & 1{,}554 & 69 & 200 & 50 \\
    Turns (decision points)          & 1{,}011 & 1{,}520 & 2{,}791 & 497 & 731 & 50 \\
    Steps (tool calls)               & 3{,}037 & 4{,}266 & 7{,}128 & 2{,}758 & 1{,}142 & 131 \\
    Turns per task (mean\,$\pm$\,sd) & 3.25\,$\pm$\,2.30 & 2.10\,$\pm$\,1.96 & 1.80\,$\pm$\,1.53
                                     & 7.20\,$\pm$\,3.66 & 3.66\,$\pm$\,1.26 & 1.00\,$\pm$\,0.00 \\
    Steps per task (mean\,$\pm$\,sd) & 9.77\,$\pm$\,5.57 & 5.88\,$\pm$\,5.26 & 4.59\,$\pm$\,4.14
                                     & 39.97\,$\pm$\,19.90 & 5.71\,$\pm$\,1.94 & 2.62\,$\pm$\,1.02 \\
    Registry size $|\reg|$           & 46 & 91 & 128 & 103 & 128 & 21 \\
    Distinct tools used              & 46 & 91 & 128 & 81 & 81 & 21 \\
    APIs for 50\%\,/\,80\% of calls  & 4\,/\,13 & 5\,/\,18 & 5\,/\,27 & 2\,/\,13 & 17\,/\,38 & 8\,/\,16 \\
    Tool-usage entropy $H$ (bits)    & 4.12 & 4.69 & 5.14 & 3.78 & 5.82 & 4.28 \\
    \quad normalized $H/\log_2\!N_{\text{used}}$ & 0.75 & 0.72 & 0.73 & 0.60 & 0.92 & 0.97 \\
    Retail\,:\,telecom (turns, real $+$ synthetic) & 727\,:\,284 & 1{,}000\,:\,520 & 1{,}935\,:\,856 & --- & --- & --- \\
    \bottomrule
  \end{tabular}}
\end{table}

\begin{multicols}{2}
\subsection{Trajectory provenance}
\label{app:prov}
Per-model turn breakdown for the retail and telecom domains
(\S\ref{sec:setup-sampling}), together with the share of each model's steps that
carry a teacher-forced reasoning trace. Retail turns are real $\tau^2$
trajectories; telecom turns are synthetic derivations
(\S\ref{sec:setup-synth}), credited to the model whose scraped submission
seeded them.

\begin{table}[H]
  \centering
  \caption{\textbf{Real $\tau^2$ retail turn provenance.}}
  \label{tab:prov-retail}
  \small
  \setlength{\tabcolsep}{3pt}
  \begin{tabular}{@{}l rrr@{}}
    \toprule
    Source model ($\tau^2$ leaderboard) & Turns & Steps & Reason.\ (\%) \\
    \midrule
    Qwen3.5-397B-A17B & 143 & 420 & 12.9 \\
    Claude Sonnet 4.5 & 142 & 307 & 50.2 \\
    GLM-5             & 142 & 389 & 67.6 \\
    Claude Opus 4.5   & 142 & 376 & 63.8 \\
    GPT-5.2           &  39 &  65 &  0.0 \\
    Gemini 3 Pro      &  18 &  34 &  0.0 \\
    Gemini 3 Flash    &  10 &  20 &  0.0 \\
    \midrule
    \textbf{Total}    & \textbf{636} & \textbf{1611} & \textbf{44.1} \\
    \bottomrule
  \end{tabular}
\end{table}

\begin{table}[H]
  \centering
  \caption{\textbf{Telecom turn provenance.} ``Authored'' are turns written
    in-house.}
  \label{tab:prov-telecom}
  \small
  \begin{tabular}{@{}l rrr@{}}
    \toprule
    Attributed model & Turns & Steps & Reason.\ (\%) \\
    \midrule
    Gemini 3 Flash    & 166 & 485 &  0.0 \\
    GPT-5.2           &  76 & 290 &  0.0 \\
    Gemini 3 Pro      &  31 &  98 &  0.0 \\
    GLM-5             &  25 & 150 & 50.0 \\
    Qwen3.5-397B-A17B &  15 &  81 & 33.3 \\
    Claude Opus 4.5   &  11 &  42 & 90.5 \\
    Claude Sonnet 4.5 &   1 &   6 & 83.3 \\
    \midrule
    Authored (in-house) & 531 & 788 & 15.5 \\
    \midrule
    \textbf{Total}    & \textbf{856} & \textbf{1940} & \textbf{13.8} \\
    \bottomrule
  \end{tabular}
\end{table}

\subsection{Curation battery prompts}
\label{app:curation}
Each gate is a topic-agnostic rubric scored by an LLM judge
(\S\ref{sec:setup-curation}): Mixed Tau2 R\&T was screened with the full battery
(\api{C1}--\api{C7} and \api{C10}, all blocking except the diagnostic \api{C3}
and the policy-level \api{C6}; per-gate rates in Table~\ref{tab:curation}),
while BFCL~v3 was screened with \api{C10} alone, a $3$-sample majority vote over
all $731$ turns against the merged $128$-tool registry that found $702$
($96.0\%$) with a unique gold tool and no combining-induced substitutes. The \api{C3} prompt itself is reproduced in \S\ref{app:prompts}.

\begin{table}[H]
  \centering
  \caption{\textbf{Curation battery.}}
  \label{tab:curation}
  \small
  \setlength{\tabcolsep}{3pt}
  \begin{tabular}{@{}ll cc@{}}
    \toprule
    Gate & What it checks & Retail & Telecom \\
    \midrule
    \api{C1} & ground-truth validity        & 1197/1299 & 647/718 \\
    \api{C2} & non-substitutability         & 1254/1299 & 327/381 \\
    \api{C3} & blind human-likeness         & 740/1299 & 495/718 \\
    \api{C4} & API disjointness             & 50/50 & 100/100 \\
    \api{C5} & non-cloning                  & 1297/1299 & 717/718 \\
    \api{C7} & reverse non-substitutability & 632/636 & 245/325 \\
    \api{C10}& real-vs-real substitutability & 633/636 & 296/325 \\
    \bottomrule
  \end{tabular}
\end{table}

\subsection{Reproduced prompts}
\label{app:prompts}
The two prompts reproduced verbatim here are the \api{C3} judge prompt, the
diagnostic that compared real $\tau^2$ trajectories against our synthetic ones
(\S\ref{sec:setup-synthval}), and the retail policy document. The
regular-split protocol replaced the single $\tau^2$-retail policy prompt with
two broadened per-domain documents, retail and telecom, routed by domain at
inference, so that no clause contradicts any gold trajectory, real or synthetic
(\S\ref{sec:setup-policy}). For space we reproduce only the retail document
below, and only in excerpt: the authentication, escalation, and order-operation
clauses, with the remainder elided at \texttt{[...]}. The telecom document is
its counterpart, broadened the same way for the telecom registry.

\end{multicols}
\paragraph{Blind real-vs-synthetic judge prompt (\api{C3}).}\mbox{}\\
\begin{tcolorbox}[breakable, colback=white, colframe=black!30, boxrule=0.4pt,
                  arc=1.5pt, left=4pt, right=4pt, top=4pt, bottom=4pt]
{\footnotesize
\begin{verbatim}
# C3 -- Real vs. synthetic (paired indistinguishability)

## SYSTEM PROMPT

---Role---
You are shown two conversations between a person and an assistant. One was written by a
human; one was written by a model. Decide which is the model-written (synthetic) one, or
that you cannot reliably tell. Judge ONLY the human-language wording.

DO:
- Judge only the dialogue's language: word choice, tone, rhythm, and how naturally the
  person and the assistant phrase things.
- Weigh natural human signals: hedging, self-corrections, brevity, small talk, the way
  real people under- or over-explain.
- If both read as equally natural language, answer "cannot_tell".

DON'T:
- Don't judge the scenario or topic. The two conversations may be about different things
  and need not line up in topic, order, or structure; any task is in scope.
- Don't judge by the number of requests, or whether they relate to each other.
- Don't use the masked tool labels (tool_a, tool_b, ...), their count, or the hidden call
  arguments/outputs -- the same redaction is applied to both sides and is not a clue.
- Don't judge by conversation length or number of turns.
- Don't reward messiness or penalize clean writing on its own, and don't try to recall or
  match specific datasets or examples.

Reason first, then output only the JSON schema below, nothing else.

---Output schema---
{
    "experiment": "C3",
    "reasoning": "<which conversation's LANGUAGE reads as less natural, and why>",
    "guess": "A" | "B" | "cannot_tell",
    "confidence": "low" | "medium" | "high"
}

## USER PROMPT

---Conversation A---
{{CONV_A}}

---Conversation B---
{{CONV_B}}

Analyze, then output the JSON.
\end{verbatim}
}
\end{tcolorbox}

\paragraph{Broadened retail policy (excerpt).}\mbox{}\\
\begin{tcolorbox}[breakable, colback=white, colframe=black!30, boxrule=0.4pt,
                  arc=1.5pt, left=4pt, right=4pt, top=4pt, bottom=4pt]
{\footnotesize
\begin{verbatim}
# Retail agent policy

[...]

You should authenticate the user identity before providing account-, order-, or
profile-specific information or taking any action on the user's account. You can
locate the user or their order by any identifier the user provides, such as their
email, their name together with their zip code, a phone number, a username, a user id,
or an order id. When the user provides a user id, username, or order id directly, you
may use it to look up the account or order without a separate identification step
first. (General product or catalog lookups that are not tied to a specific account may
be handled without prior authentication.)

[...]

You should transfer the user to a human agent if and only if the request cannot be
handled within the scope of your actions. To transfer, first make a tool call to
transfer_to_human_agents, and then send the message 'YOU ARE BEING TRANSFERRED TO A
HUMAN AGENT. PLEASE HOLD ON.' to the user.

[...]

## Generic action rules

You do not need to look up or re-verify an order's status before choosing a
tool; select the operation that matches the user's request, even when the order is at
an intermediate status or no status has been retrieved yet.

When exchanging or modifying items, be sure that all items to be changed are collected
into a list before making the tool call. A given exchange or item-modification action
is generally applied once per order.

## Cancel pending order

The cancel-pending-order action is the tool for cancelling an order that is still
pending (before it ships). Cancelling a delivered order, or cancelling an individual
item, are separate operations with their own dedicated tools and are not limited to
pending orders.

The user needs to confirm the order id and the reason for cancellation (commonly
'no longer needed' or 'ordered by mistake').

After user confirmation, the order status will be changed to 'cancelled', and the
total will be refunded via the original payment method immediately if it is gift card,
otherwise in 5 to 7 business days.

## Modify pending order

The modify-pending-order actions are the tools for changing an order that is still
open -- typically a pending order, including one already altered by an earlier change
(for example a status such as 'pending (item modified)', 'address_updated', or
'items_modified').

For a pending order, the modify-pending-order actions cover its shipping address,
payment method, or product item options. A pending order can also be the subject of
other order, fulfillment, and post-purchase services (for example a shipping upgrade,
delivery scheduling, a split shipment, item-level cancellation, or applying a gift
card or discount) through the appropriate dedicated tool.

### Modify payment

The user can only choose a single payment method different from the original payment
method.

If the user wants the modify the payment method to gift card, it must have enough
balance to cover the total amount.

After user confirmation, the order status will be kept as 'pending'. The original
payment method will be refunded immediately if it is a gift card, otherwise it will be
refunded within 5 to 7 business days.

### Modify items

This action changes the order status to 'pending (item modified)' and finalizes the
item modifications. So you must confirm all the details are correct and be cautious
before taking this action. In particular, remember to remind the customer to confirm
they have provided all the items they want to modify.

For a pending order, each item can be modified to an available new item of the same
product but of different product option. There cannot be any change of product types,
e.g. modify shirt to shoe.

The user must provide a payment method to pay or receive refund of the price
difference. If the user provides a gift card, it must have enough balance to cover the
price difference.

## Return delivered order

The return-delivered-order action is the tool for returning items from an order that
has been delivered. Returning items on a pending order is a separate operation with
its own dedicated tool.

The user needs to confirm the order id and the list of items to be returned.

The user needs to provide a payment method to receive the refund.

The refund must either go to the original payment method, or an existing gift card.

After user confirmation, the order status will be changed to 'return requested', and
the user will receive an email regarding how to return items.

## Exchange delivered order

The exchange-delivered-order action is the tool for exchanging items from a delivered
order. In particular, remember to remind the customer to confirm they have provided
all items to be exchanged.

For a delivered order, an item may be exchanged for an available new item -- commonly
the same product with a different product option, and in some cases a different
product.

The user must provide a payment method to pay or receive refund of the price
difference. If the user provides a gift card, it must have enough balance to cover the
price difference.

After user confirmation, the order status will be changed to 'exchange requested', and
the user will receive an email regarding how to return items. There is no need to place
a new order.
\end{verbatim}
}
\end{tcolorbox}

\begin{multicols}{2}
\subsection{Example synthetic trajectories}
\label{app:synthtraj}
The blind indistinguishability gate (\api{C3}, Table~\ref{tab:curation})
places a synthetic trajectory next to the genuine $\tau^2$-bench conversation
it was judged against. Figure~\ref{fig:fooled} is one pair from the blind human
study (\S\ref{sec:setup-human}); \textbf{A} is the assistant and \textbf{U} the
user.

\begin{figure}[H]
  \centering
  \begin{tikzpicture}[
    font=\scriptsize,
    bub/.style={rounded corners=3pt, inner sep=3pt, align=left,
                text width=0.90\linewidth, line width=0.4pt},
    ag/.style={bub, fill=black!5, draw=black!18},
    us/.style={bub, fill=deep!8, draw=deep!30},
    hd/.style={font=\scriptsize\bfseries, align=center, inner sep=2pt,
               text width=0.90\linewidth, rounded corners=2pt, fill=black!10},
    note/.style={font=\scriptsize\itshape, text=black!55, align=left,
                 text width=0.96\linewidth},
  ]
  \node[hd] (h2) at (0,0) {$\tau^2$-bench (real reference)};
  \node[us, below=3pt of h2.south west, anchor=north west] (r1)
    {\textbf{U}\ \ Quick fun thing first: can you guess a famous poem?
     ``Two roads diverged in a yellow wood\ldots''};
  \node[ag, below=3pt of r1.south west, anchor=north west] (r2)
    {\textbf{A}\ \ That's ``The Road Not Taken'' by Robert Frost---a classic!
     However, I'm a customer service agent here to help with orders, returns,
     and account information\ldots};
  \node[us, below=3pt of r2.south west, anchor=north west] (r3)
    {\textbf{U}\ \ I placed a recent order and I'm trying to check one item: the
     Bluetooth speaker. Can you tell me the price and its battery life? Also,
     the order should be under Chen Johnson (Houston, TX 77004).};

  \node[hd, below=6pt of r3.south west, anchor=north west] (h1)
    {Ours (synthetic)};
  \node[us, below=3pt of h1.south west, anchor=north west] (l1)
    {\textbf{U}\ \ Hi, I was hoping to get some help with my account today.};
  \node[ag, below=3pt of l1.south west, anchor=north west] (l2)
    {\textbf{A}\ \ Could you confirm your name and ZIP code?};
  \node[us, below=3pt of l2.south west, anchor=north west] (l3)
    {\textbf{U}\ \ Yeah, sure, it's Nico Ferri, ZIP 02139. My order \#W3410700
     came in and I want to send part of it back---could you get me a printable
     return label?};

  \node[note, below=4pt of l3.south west, anchor=north west] {Tool calls are
    masked on both sides and each runs to the current turn. The reviewer and
    the \api{C3} judge both chose the lower conversation as human-authored.};
  \end{tikzpicture}
  \caption{\textbf{A synthetic trajectory humans read as real.}}
  \label{fig:fooled}
\end{figure}

Our synthetic user turns are short, colloquial, and carry one clear intent
(``Yeah, sure, it's Nico Ferri\ldots could you get me a printable return
label''); the genuine $\tau^2$-bench side more often reads as generated, with
meta assistant boilerplate and long, enumerated requests. That difference in
\emph{register}, not domain content, is what the battery preserves, and why
neither human nor model reviewers reliably recover the $\tau^2$-bench
conversation.

\subsection{Human study protocol}
\label{app:humanstudy}
The three reviewers of \S\ref{sec:setup-human} were given the \api{C3} judge's
criteria (\S\ref{app:prompts}) through a review portal, which asked which
conversation was human-authored and took a confidence rating; pairs answered
``not sure'' are the undecided ones excluded from the $59$. They were colleagues
at our institution, unpaid, uninvolved in the project, not reporting to any
author, and free to decline. They were not told beforehand that this was a
research study, since disclosure could bias the judgment; all three were
debriefed afterward and consented to our use of their judgments. We collected no
demographics, and the protocol was not submitted to an ethics review board: no
personal data was collected and the task was to rate fictional dialogue.

\subsection{Benchmark adaptations}
\label{app:adapt}

\begin{table}[H]
  \centering
  \caption{\textbf{Our setting against the published benchmarks.}}
  \label{tab:adapt}
  \small
  \setlength{\tabcolsep}{3pt}
  \begin{tabular}{@{}ll >{\raggedright\arraybackslash}p{3.5cm}@{}}
    \toprule
    Aspect & Where & Original $\to$ ours \\
    \midrule
    Registry & Mixed Tau2 R\&T & domains separate $\to$ pooled, three sizes \\
    Tools    & Mixed Tau2 R\&T & real only $\to$ plus validated synthetic \\
    Backends & BFCL~v3 & four, separate $\to$ merged, alternates checked \\
    Policy   & all & absent or $\tau^2$ $\to$ $\tau^2$-style, ours \\
    Grading  & all & per task $\to$ per turn \\
    \bottomrule
  \end{tabular}
\end{table}


\subsection{Method details}
\label{app:method}

\paragraph{Weight model and training.}
The softplus floor is $\varepsilon{=}10^{-3}$. Parameters
$(\bm a,b_w,\bm c,b_T)$ are fit with Adam (learning rate $0.05$, $500$ epochs,
$\beta_1{=}0.9$, $\beta_2{=}0.999$, $\epsilon_{\text{Adam}}{=}10^{-8}$), an
$\ell_2$ penalty of $10^{-3}$ on the slope vectors $\bm a,\bm c$ only (biases
unregularized), and seed $0$. With $t^\star$ the gold child of
a training step, $j$ the trajectory that step came from, and $\mathcal{B}$ the
set of such steps drawn from $\mathcal{D}$, the objective is
\begin{align}
\mathcal{L}(\bm\theta)
&= -\frac{1}{|\mathcal{B}|}\!\!\sum_{(q,S_i,t^\star,j)\in\mathcal{B}}\!\!
\log P^{\setminus j}_{\bm\theta}\!\big(t^\star\mid q,S_i\big),\notag\\[2pt]
P^{\setminus j}_{\bm\theta}
&=\softmax\nolimits_{t\in\children^{\setminus j}(S_i)}\; s^{\setminus j}(t).
\label{eq:nll}
\end{align}
where the superscript $\setminus j$ marks a quantity computed against a trie and
edge memory built without trajectory $j$, over five cross-fit folds.

\columnbreak

\par\addvspace{\intextsep}
\hrule height.8pt depth0pt \kern2pt
{\captionsetup{style=ruled}%
 \captionof{algorithm}{One \tracer{} retrieval step at state $S_i$ for request $q$.}%
 \label{alg:step}}
\kern2pt\hrule\kern2pt
\nopagebreak
\begin{algorithmic}[1]
\Require trie with edge counts; edge memory $m$; blend heads $(\bm a,b_w,\bm c,b_T)$
  and standardizer $(\bm\mu,\bm\sigma)$; embedder $e$; gate $\tau$; budget $k$;
  optional global set $G$
\State $\children \gets \textsc{Children}(S_i)$
\If{$\children = \varnothing$}
  \State $\mathcal{K} \gets \textsc{GlobalVectorSearch}(q,k)$
\Else
  \State $P(t) \gets \log\big(\Ptrie(t\mid S_i)+\epsilon\big)$
  \State $R(t) \gets \cossim\!\big(e(\ctx(q,S_i)),\,m(S_i\!\to\!t)\big)$
  \State $L(t) \gets \log \normR(t)$
  \State $\bm z \gets (\bm x(q,S_i)-\bm\mu)\,/\,\bm\sigma$
  \State $w \gets \sigma(\bm a^{\!\top}\bm z + b_w)$
  \State $T \gets \softplus(\bm c^{\!\top}\bm z + b_T)+\varepsilon$
  \State $s(t) \gets \tfrac{1}{T}\big[(1-w)L(t) + w\,P(t)\big]$
  \State $\bm p \gets \softmax_{t\in\children} s(t)$;\quad $\kappa \gets \max_t p(t)$
  \If{$\kappa < \tau$}
    \State $\mathcal{K} \gets \textsc{GlobalVectorSearch}(q,k)$
  \Else
    \State $\mathcal{K} \gets \textsc{TopK}(s,\,k)$
  \EndIf
\EndIf
\If{extension enabled}
  \State $\mathcal{K} \gets \mathcal{K}\,\cup\,G$
\EndIf
\State \Return $\mathcal{K}$
\end{algorithmic}
\kern2pt\hrule
\par\addvspace{\intextsep}

\end{multicols}

\begin{figure}[t]
  \centering
  \resizebox{\linewidth}{!}{%
  \begin{tikzpicture}[
    font=\small,
    bx/.style={rounded corners=2.5pt, draw, line width=0.7pt, align=center,
               inner sep=4pt, minimum height=9mm, minimum width=21mm},
    inb/.style={bx, draw=vizorange!75!black, fill=vizorange!10},
    selb/.style={bx, draw=vizred!75!black,    fill=vizred!9},
    fbb/.style={bx, draw=vizblue!75!black,   fill=vizblue!10},
    outb/.style={bx, draw=vizaqua!65!black,  fill=vizaqua!12},
    ar/.style={-{Stealth[length=5pt,width=4pt]}, line width=0.7pt, draw=black!62},
    lbl/.style={font=\scriptsize\itshape, text=black!60},
    bandlbl/.style={font=\small\bfseries},
    tn/.style={rounded corners=2pt, draw=black!45, fill=white, line width=0.5pt,
               font=\ttfamily\scriptsize, inner sep=2.5pt, minimum height=5mm},
    tnhi/.style={tn, draw=vizred!75!black, fill=vizred!16, line width=1pt},
    tncand/.style={tn, draw=vizred!55!black, fill=vizred!5},
    tnfade/.style={tn, draw=black!22, text=black!42},
    ted/.style={draw=black!45, line width=0.6pt},
    tedhi/.style={draw=vizred!70!black, line width=1pt},
    elbl/.style={font=\scriptsize, text=black!55, inner sep=1.5pt, fill=white},
  ]

  \begin{scope}[shift={(0,7.75)}, every node/.append style={font=\ttfamily\scriptsize}]
    \node[tn] (root) at (9.2, 0) {root};

    \node[tn]     (a1) at ( 2.3,-1.15) {find\_user\_id\_by\_email};
    \node[tn]     (a2) at ( 8.0,-1.15) {find\_user\_id\_by\_name\_zip};
    \node[tnfade] (a3) at (13.2,-1.15) {get\_product\_details};
    \node[tnfade] (a4) at (17.3,-1.15) {\dots\ 13 more};

    \node[tn]     (b1) at ( 2.3,-2.30) {get\_user\_details};
    \node[tnfade] (b2) at ( 6.1,-2.30) {get\_order\_details};
    \node[tnfade] (b3) at ( 9.4,-2.30) {get\_user\_details};
    \node[tnfade] (b4) at (12.6,-2.30) {list\_all\_product\_types};
    \node[tnfade] (b5) at (16.6,-2.30) {\dots};

    \node[tnhi]   (c1) at ( 3.1,-3.45) {get\_order\_details};
    \node[tnfade] (c2) at ( 7.3,-3.45) {get\_order\_status};
    \node[tnfade] (c3) at (10.6,-3.45) {\dots};

    \node[tncand] (d1) at ( 0.0,-4.60) {get\_return\_label};
    \node[tncand] (d2) at ( 3.4,-4.60) {split\_order\_shipment};
    \node[tncand] (d3) at ( 7.0,-4.60) {add\_gift\_message};

    \draw[ted]          (root) -- node[elbl,pos=0.62]{$0.22$} (a1);
    \draw[ted]          (root) -- node[elbl,pos=0.62]{$0.18$} (a2);
    \draw[ted,black!25] (root) -- node[elbl,pos=0.62]{$0.11$} (a3);
    \draw[ted,black!20] (root) -- (a4);
    \draw[ted]          (a1) -- node[elbl,pos=0.5]{$0.88$} (b1);
    \draw[ted,black!25] (a1) -- node[elbl,pos=0.55]{$0.12$} (b2);
    \draw[ted,black!25] (a2) -- (b3);
    \draw[ted,black!20] (a3) -- (b4);
    \draw[ted,black!20] (a3) -- (b5);
    \draw[ted]          (b1) -- node[elbl,pos=0.5]{$0.71$} (c1);
    \draw[ted,black!25] (b1) -- node[elbl,pos=0.55]{$0.19$} (c2);
    \draw[ted,black!20] (b3) -- (c3);
    \draw[tedhi]        (c1) -- node[elbl,pos=0.55]{$0.41$} (d1);
    \draw[tedhi]        (c1) -- node[elbl,pos=0.55]{$0.22$} (d2);
    \draw[tedhi]        (c1) -- node[elbl,pos=0.55]{$0.15$} (d3);

    \node[lbl,anchor=east,text=vizred!70!black] at ([xshift=-2mm]c1.west)
      {current state $S_i$};
    \node[lbl,anchor=west,text=vizred!60!black] at ([xshift=2mm]d3.east)
      {$\children(S_i)$: the only tools ranked};
    \node[lbl,anchor=west] at ([xshift=2mm]a4.east)
      {edge label $=\Ptrie(t\mid S_i)$, the \emph{P} signal;};
    \node[lbl,anchor=west] at ([xshift=2mm,yshift=-4mm]a4.east)
      {each edge also stores $m(S_i\!\to\!t)\Rightarrow$ the \emph{L} signal};
  \end{scope}

  \begin{scope}[on background layer]
    \node[fit=(root)(a4)(d1)(d3), rounded corners=4pt, draw=black!22,
          fill=black!2, inner sep=7pt] (bandtrie) {};
  \end{scope}
  \node[bandlbl, text=black!55, anchor=south west] at (bandtrie.north west)
    {The execution trie, built from training trajectories};

  \node[inb] (q) at (0,0.95)  {User request $q$};
  \node[inb] (s) at (0,-0.95) {Turn prefix $S_i$\\[-1pt]{\scriptsize$(t_1,\dots,t_i)$}};

  \begin{scope}[shift={(3.55,-0.40)}]
    \node[circle,draw=black!45,fill=white,inner sep=1.3pt]              (n0) at (0,1.15) {};
    \node[circle,draw=black!45,fill=white,inner sep=1.3pt]              (n1) at (-0.55,0.42) {};
    \node[circle,draw=vizred!75!black,fill=vizred!55,inner sep=2.0pt]   (n2) at (0.45,0.42) {};
    \node[circle,draw=vizred!70!black,fill=white,inner sep=1.3pt]       (n3) at (0.00,-0.34) {};
    \node[circle,draw=vizred!70!black,fill=white,inner sep=1.3pt]       (n4) at (0.45,-0.34) {};
    \node[circle,draw=vizred!70!black,fill=white,inner sep=1.3pt]       (n5) at (0.90,-0.34) {};
    \draw[draw=black!42,line width=0.5pt] (n0)--(n1) (n0)--(n2);
    \draw[draw=vizred!70!black,line width=0.6pt] (n2)--(n3) (n2)--(n4) (n2)--(n5);
    \node[lbl,anchor=south]      at (0.20,1.30)  {execution trie};
    \node[lbl,anchor=east,text=vizred!70!black] at (0.30,0.42) {$S_i$};
    \node[lbl,anchor=north]      at (0.45,-0.52) {$\children(S_i)$};
  \end{scope}
  \coordinate (jn) at (1.75,0);

  \node[selb, minimum height=11mm] (p) at (6.5, 1.05)
        {Empirical prior\\[-1pt]$P(t)=\log\Ptrie(t\mid S_i)$};
  \node[selb, minimum height=11mm] (l) at (6.5,-1.05)
        {Contextual prior\\[-1pt]$L(t)=\log\normR(t)$};
  \node[selb] (b) at (9.9, 0)    {Learned blend\\[-1pt]$(w,T)$ from $\bm x$};
  \node[selb] (sc) at (13.0,0)   {$s(t)=\frac{1}{T}\big[(1{-}w)L{+}wP\big]$\\[-1pt]
                                  {\scriptsize rank, take top-$k$}};

  \node[fbb] (fb) at (13.0,-2.55) {Global vector search\\[-1pt]{\scriptsize
      unseen state, or $\kappa<\tau$}};

  \node[outb] (K)  at (17.6, 0.95) {Candidate set $\mathcal{K}$\\[-1pt]{\scriptsize$k\ll|\reg|$}};
  \node[outb] (ag) at (17.6,-0.95) {Agent};
  \node[outb] (tc) at (21.5,-0.95) {Next tool call};

  \begin{scope}[on background layer]
    \node[fit=(q)(s), rounded corners=4pt, draw=vizorange!45,
          fill=vizorange!5, inner sep=6pt] (bandin) {};
    \node[fit=(n0)(n5)(p)(l)(b)(sc), rounded corners=4pt, draw=vizred!40,
          fill=vizred!4, inner sep=7pt] (bandsel) {};
    \node[fit=(K)(ag), rounded corners=4pt, draw=vizaqua!45,
          fill=vizaqua!5, inner sep=6pt] (bandout) {};
  \end{scope}

  \node[bandlbl, text=vizorange!70!black, anchor=north] at (bandin.south)   {(1) Inputs};
  \node[bandlbl, text=vizred!70!black,    anchor=north west]
        at (bandsel.south west) {(2) State-conditioned Selector};
  \node[bandlbl, text=vizaqua!55!black,   anchor=north] at (bandout.south)  {(4) Agent};
  \node[bandlbl, text=vizblue!70!black, anchor=north]
        at ([yshift=-1mm]fb.south) {(3) OOD fallback};

  \coordinate (gx) at (3.9,0);
  \draw[-{Stealth[length=6pt,width=5pt]}, line width=1pt, draw=vizred!60!black]
        (bandtrie.south -| gx) -- ++(0,-0.62)
        node[midway, right, font=\scriptsize\itshape, text=vizred!65!black]
        {\ consulted at every step};
  \draw[ar] (q.east) -| (jn);
  \draw[ar] (s.east) -| (jn);
  \draw[ar] (jn) -- (n0);
  \draw[ar] (n4.east) to[out=0,in=180] (p.west);
  \draw[ar] (n5.east) to[out=0,in=180] (l.west);
  \draw[ar] (p.east) -- ++(0.45,0) |- (b.west);
  \draw[ar] (l.east) -- ++(0.45,0) |- (b.west);
  \draw[ar] (b) -- (sc);
  \draw[ar] (sc.south) -- (fb.north);
  \coordinate (bus) at ($(K.west)+(-0.55,0)$);
  \draw[line width=0.7pt, draw=black!62] (sc.east) -- (sc.east -| bus);
  \draw[line width=0.7pt, draw=black!62] (fb.east) -- (fb.east -| bus);
  \draw[line width=0.7pt, draw=black!62] (fb.east -| bus) -- (bus);
  \draw[ar] (bus) -- (K.west);
  \draw[ar] (K) -- (ag);
  \draw[ar] (ag) -- (tc);
  \end{tikzpicture}}
  \caption{\textbf{Full \tracer{} architecture.}}
  \label{fig:arch}
\end{figure}

\begin{multicols}{2}

Since $1/T$ is a common positive factor in Eq.~\eqref{eq:headline}, the ranking
depends only on $w$; $T$ sets the sharpness of $\softmax_{t}s(t)$, whose maximum
$\kappa$ is the retrieval confidence read by the gate (\S\ref{sec:gate}).

\paragraph{Selecting the global set.}
The prefix entropy of \S\ref{sec:setup-baselines}, in bits, is
\[
  H(S\mid t)\;=\;-\!\!\sum_{S\in\mathcal{S}(t)}\!\pi(S\mid t)\,\log_2 \pi(S\mid t),
\]
where $\mathcal{S}(t)=\{S: t\in\children(S)\}$ and
$\pi(S\mid t)=\#(S\!\to\!t)/\sum_{S'\in\mathcal{S}(t)}\#(S'\!\to\!t)$. Ties are
broken by tool name.

\subsection{Feature selection and weight-head ablations}
\label{app:features}

The five-feature weight head of \S\ref{sec:wt}: which signals it uses, why
those five, and what is lost by giving the head more capacity.

\paragraph{Feature pool and selection.}
The head uses five interpretable signals drawn from a fourteen-feature pool of
\emph{prior} statistics (n\_children, prior\_entropy, prior\_top1,
prior\_margin, log\_node\_count, depth), \emph{likelihood} statistics
(R\_top1, R\_margin, R\_mean, R\_std, top\_edge\_support), and \emph{context}
statistics (n\_prior\_turns, n\_prior\_calls, query\_len).
The five span all three groups while avoiding the strong within-group
collinearity of the pool (e.g.\ prior\_top1 vs.\
prior\_margin). Writing $p_{(1)}\ge p_{(2)}$ for the two largest
$\Ptrie(\cdot\mid S_i)$, $\#(S_i\!\to\!t)$ for the number of training turns that
took $t$ from $S_i$, and $\hat t=\arg\max_{t} R(t)$, they are
{\small\begin{align*}
x_1=\text{prior\_margin}      &= p_{(1)}-p_{(2)},\\
x_2=\text{top\_edge\_support} &= \log\max\!\big(1,\#(S_i\!\to\!\hat t)\big),\\
x_3=\text{R\_top1}            &= \max\nolimits_{t\in\children(S_i)} R(t),\\
x_4=\text{query\_len}         &= \text{words in the user request},\\
x_5=\text{n\_prior\_turns}    &= \text{number of prior turns},
\end{align*}}
covering \emph{prior confidence} ($x_1$), \emph{edge-memory evidence}
($x_2,x_3$), and \emph{context scale} ($x_4,x_5$). Each is standardized with the
mean and standard deviation of the training folds only, frozen at inference.

\paragraph{Weight-head capacity ablation.}
Under an identical split grouped by task, a PCA-5 linear head and a $\tanh$ MLP
($16$ hidden units, learning rate $0.01$, $1500$ epochs) stay within $0.3$
points of the hand-picked linear head, which itself beats a constant
$w{=}0.5$ by $18$ points (Table~\ref{tab:headcap}).

\begin{table}[H]
  \centering
  \caption{\textbf{Weight-head capacity.} Mixed Tau2 R\&T, five-fold
    \textsc{GroupKFold}, $1{,}576$ steps.}
  \label{tab:headcap}
  \small
  \begin{tabular}{@{}l c@{}}
    \toprule
    Weight head & Recall@5 (\%) \\
    \midrule
    Constant $w{=}0.5$ (baseline)     & 80.3 \\
    Hand-picked linear (5 features)   & 98.4 \\
    PCA-5 linear (of 14)              & 98.6 \\
    $\tanh$ MLP, hidden $16$          & 98.5 \\
    \bottomrule
  \end{tabular}
\end{table}

\subsection{What the head learns}
\label{app:learned}

Table~\ref{tab:wbyenv} reports the mean and across-fold SD of the per-step
weight the head predicts in each environment. $w$ weights the empirical prior in
Eq.~(\ref{eq:headline}), so small $w$ means the contextual prior carries the
ranking. Free-running was not run on Tau3~Banking or API-Bank (n/a).

\begin{table}[H]
  \centering
  \caption{\textbf{Learned blend weight $w$ by environment.}}
  \label{tab:wbyenv}
  \small
  \resizebox{\linewidth}{!}{%
  \begin{tabular}{@{}l cc@{}}
    \toprule
    Environment & Teacher-forcing & Free-running \\
    \midrule
    Mixed Tau2 R\&T (Small)      & $0.014\pm0.004$ & $0.012\pm0.002$ \\
    Mixed Tau2 R\&T (Medium)     & $0.014\pm0.003$ & $0.015\pm0.000$ \\
    Mixed Tau2 R\&T (Large)      & $0.019\pm0.002$ & $0.022\pm0.001$ \\
    Tau3 Banking                 & $0.167\pm0.037$ & n/a \\
    API-Bank                     & $0.255\pm0.086$ & n/a \\
    BFCL v3                      & $0.024\pm0.015$ & $0.024\pm0.015$ \\
    \bottomrule
  \end{tabular}}
\end{table}

Table~\ref{tab:wcoef} reports the fold mean and SD of each coefficient of the
logistic head that predicts $w$, at the three registry sizes. Features are
standardised, so the coefficients are comparable within a row. \tracer{} and
\tracers{} share a scorer, and their fits coincide to within $0.006$ on every
feature.

\end{multicols}
\begin{table}[H]
  \centering
  \caption{\textbf{Learned $w$ coefficients, Mixed Tau2 R\&T.}}
  \label{tab:wcoef}
  \scriptsize
  \setlength{\tabcolsep}{3pt}
  \begin{tabular}{@{}ll cccccc@{}}
    \toprule
    Reg. & Alg. & \api{prior\_margin} & \api{top\_edge\_support} & \api{R\_top1} & \api{query\_len} & \api{n\_prior\_turns} & $b_w$ \\
    \midrule
    Small   & \tracer{}    & $0.544\pm0.131$ & $-0.225\pm0.110$ & $-0.298\pm0.127$ & $0.181\pm0.044$ & $-0.562\pm0.166$ & $-4.337\pm0.336$ \\
    Medium  & \tracer{}    & $0.120\pm0.165$ & $-0.417\pm0.010$ & $-0.216\pm0.099$ & $0.124\pm0.035$ & $-0.675\pm0.162$ & $-4.284\pm0.225$ \\
    Large   & \tracer{}    & $0.051\pm0.070$ & $-0.364\pm0.113$ & $-0.139\pm0.036$ & $0.118\pm0.029$ & $-0.511\pm0.083$ & $-3.940\pm0.113$ \\
    \midrule
    Small   & \tracers{}   & $0.543\pm0.132$ & $-0.225\pm0.110$ & $-0.298\pm0.128$ & $0.181\pm0.044$ & $-0.562\pm0.166$ & $-4.337\pm0.338$ \\
    Medium  & \tracers{}   & $0.120\pm0.166$ & $-0.418\pm0.011$ & $-0.216\pm0.099$ & $0.124\pm0.035$ & $-0.675\pm0.162$ & $-4.284\pm0.225$ \\
    Large   & \tracers{}   & $0.054\pm0.073$ & $-0.358\pm0.122$ & $-0.138\pm0.037$ & $0.118\pm0.028$ & $-0.512\pm0.085$ & $-3.938\pm0.116$ \\
    \bottomrule
  \end{tabular}
\end{table}

\begin{multicols}{2}
\subsection{Trie key depth}
\label{app:last2}

\tracer{} keys the trie on the full current-turn prefix (\S\ref{sec:trie}).
Table~\ref{tab:last2} varies that one choice on Mixed Tau2 R\&T (Large) with
Grok-4-20-r under teacher forcing, five folds: the last-2 index keys on the two
most recent calls, while the edge memories still mean-pool the whole
conversation and the learned blend (\S\ref{sec:wt}) is unchanged. Parentheses
give the across-fold SD.
The two keys are indistinguishable on the steps that do fall outside the trie,
so the difference is coverage rather than ranking.

\begin{table}[H]
  \centering
  \caption{\textbf{Full prefix against a two-step trie key.}}
  \label{tab:last2}
  \small
  \setlength{\tabcolsep}{4pt}
  \begin{tabular}{@{}l ccc ccc@{}}
    \toprule
    & \multicolumn{3}{c}{Selection accuracy}
    & \multicolumn{3}{c}{Out of trie} \\
    \cmidrule(lr){2-4}\cmidrule(lr){5-7}
    $k$ & Last-2 & Full & $\Delta$ & Last-2 & Full & $\Delta$ acc \\
    \midrule
    $5$  & 61.6 (1.9) & 59.5 (3.8) & $+2.1$ & $9.6\%$ & $12.4\%$ & $+1.7$ \\
    $10$ & 61.2 (1.8) & 58.3 (3.8) & $+2.9$ & $9.6\%$ & $12.4\%$ & $-0.2$ \\
    $20$ & 60.4 (2.2) & 58.5 (3.7) & $+1.9$ & $9.6\%$ & $12.4\%$ & $+0.9$ \\
    $30$ & 60.2 (2.0) & 58.6 (4.3) & $+1.5$ & $9.6\%$ & $12.4\%$ & $-0.2$ \\
    \bottomrule
  \end{tabular}
\end{table}

\subsection{Qualitative examples}
\label{app:cases}

Figure~\ref{fig:overreach} is the retail over-reach summarized in
\S\ref{sec:results-why}. The customer asks only for a prepaid return label;
\llmonly{} performs the return itself as a prerequisite, an irreversible
action, while the trie has no observed transition to that tool from this state
and leaves it outside the candidate set.

\end{multicols}
\begin{figure}[H]
  \centering
  \renewcommand{\api}[1]{\texttt{#1}}%
  \begin{tikzpicture}[
    font=\scriptsize,
    bub/.style={rounded corners=3pt, inner sep=4pt, align=left, line width=0.4pt},
    usr/.style={bub, fill=vizred!8,    draw=vizred!35},
    pre/.style={bub, fill=black!3,     draw=black!18},
    fe/.style ={bub, fill=black!4,     draw=black!30, dashed},
    agt/.style={bub, fill=vizblue!8,   draw=vizblue!40},
    cal/.style={bub, fill=vizorange!12,draw=vizorange!55},
    tag/.style={font=\tiny\itshape, text=vizmuted, inner sep=1pt},
  ]
    \node[usr, text width=0.86\linewidth, anchor=north] (cust) at (0,0)
      {\textbf{Customer:} ``Review my two recent orders and generate a prepaid
       return label for \#W3568250---it's the wrong size. My email is
       \texttt{zane.vin@example.com}.''};
    \node[pre, text width=0.86\linewidth, anchor=north west]
      at ($(cust.south west)+(0,-0.10)$) (pfx)
      {\emph{Steps 0--3 agree:} \api{find\_user\_id\_by\_email} $\to$
       \api{get\_user\_details} $\to$ \api{get\_order\_details} $\to$
       \api{get\_order\_details}};

    \coordinate (L) at ($(pfx.south west)+(0,-0.42)$);
    \coordinate (R) at ($(pfx.south west)+(0.455\linewidth,-0.42)$);

    \node[font=\scriptsize\bfseries, anchor=north west] (lh) at (L) {\tracer{}};
    \node[fe,  text width=0.40\linewidth, anchor=north west]
      at ($(lh.south west)+(0,-0.06)$) (lf)
      {\textbf{front-end.} \api{return\_\allowbreak delivered\_\allowbreak order\_\allowbreak items}: $9$th by
       contextual prior, empirical prior $0$ from this state
       $\Rightarrow$ outside the top~$5$. \api{get\_return\_label} presented.};
    \node[agt, text width=0.40\linewidth, anchor=north west]
      at ($(lf.south west)+(0,-0.08)$) (la)
      {\textbf{back-end.} ``fetch the return label for the specified delivered
       order''};
    \node[cal, text width=0.40\linewidth, anchor=north west]
      at ($(la.south west)+(0,-0.08)$) (lc)
      {\api{get\_return\_label(\ldots)}~\cmark};

    \node[font=\scriptsize\bfseries, anchor=north west] (rh) at (R) {\llmonly{}};
    \node[fe,  text width=0.40\linewidth, anchor=north west]
      at ($(rh.south west)+(0,-0.06)$) (rf)
      {\textbf{front-end.} None. All $128$ tools are placed in context,
       \api{return\_\allowbreak delivered\_\allowbreak order\_\allowbreak items} among them.};
    \node[agt, text width=0.40\linewidth, anchor=north west]
      at ($(rf.south west)+(0,-0.08)$) (ra)
      {\textbf{back-end.} ``initiate the return for order \#W3568250 so a return
       label can be generated afterward''};
    \node[cal, text width=0.40\linewidth, anchor=north west]
      at ($(ra.south west)+(0,-0.08)$) (rc)
      {\api{return\_\allowbreak delivered\_\allowbreak order\_\allowbreak items(\ldots)}~\xmark};

    \node[fit=(lh)(lc)(rh)(rc), inner sep=0pt] (cols) {};
    \draw[black!30, dashed, line width=0.4pt]
      ($(cols.north)+(0,0.12)$) -- ($(cols.south)+(0,-0.06)$);
  \end{tikzpicture}
  \caption{\textbf{The trie blocks an over-reach} (Mixed Tau2 R\&T Large,
    GPT-5.4, $k{=}5$).}
  \label{fig:overreach}
\end{figure}

\begin{multicols}{2}
\subsection{Compute, software, and artifacts}
\label{app:compute}

\paragraph{Model size and compute.}
The only trained component is the blend head of \S\ref{sec:wt}: two linear
heads over the five standardized features of \S\ref{app:features}, so
$\bm\theta=(\bm a,b_w,\bm c,b_T)$ holds $12$ parameters. It fits on CPU on one
workstation VM; no GPU was used at any point, so we report no GPU hours. All
inference ran on hosted endpoints through an internal service that does not
expose per-call cost, so we report no dollar figure. The embedder and the four
evaluated models are proprietary, with undisclosed parameter counts. Inference
spans four models and five algorithms over five folds in six environments at
each swept budget (\S\ref{sec:setup-protocol}).

\paragraph{Software and settings.}
We use NumPy, scikit-learn (\textsc{GroupKFold} for the task-grouped folds),
PyTorch for the blend head, \api{tiktoken} for token counts, and the
provider's Python client for hosted inference. Decoding uses temperature $1$,
other parameters at provider defaults. The hyperparameters in
\S\ref{app:method} and \S\ref{app:features} were hand-picked; we ran no search.

\paragraph{Artifacts and licenses.}
$\tau^2$-bench and $\tau^3$-Knowledge are MIT licensed, BFCL~v3 is Apache-2.0,
and API-Bank is MIT; \api{text-embedding-3-small} is used under its provider's
terms of service. All six environments are English-language. We use each to
evaluate tool-using agents, the purpose each was released for. We release no
artifacts here; the synthetic trajectories and
broadened policy documents we build derive from $\tau$-bench leaderboard
submissions and MIT-licensed benchmark data, and any later release will carry a
compatible license.

\paragraph{Risks.}
\tracer{} narrows the tools an agent sees, so a retrieval failure can hide a
needed tool; the confidence gate and out-of-trie fallback (\S\ref{sec:gate})
bound that, and Table~\ref{tab:ood} quantifies where it still fails. Built from
observed trajectories, the trie inherits their biases and will under-serve rare
but legitimate tool sequences. We see no societal risk beyond those already
present in tool-using agents.

\subsection{Out-of-trie coverage}
\label{app:ood}

How often a step has no matching prefix node, and how much worse selection is
on those steps (\S\ref{sec:results-fr}).

\end{multicols}
\begin{table}[H]
  \centering
  \caption{\textbf{Out-of-trie steps: how often, and how much worse.}}
  \label{tab:ood}
  \footnotesize
  \setlength{\tabcolsep}{3pt}
  \begin{tabular}{@{}ll cccccc ccc@{}}
    \toprule
    & & \multicolumn{6}{c}{Teacher-forcing}
    & \multicolumn{3}{c}{Free-running} \\
    \cmidrule(lr){3-8}\cmidrule(lr){9-11}
    & & \multicolumn{3}{c}{Mixed Tau2 R\&T}
    & \shortstack{Tau3\\Banking} & \shortstack{API-\\Bank}
    & \shortstack{BFCL\\v3}
    & \multicolumn{3}{c}{Mixed Tau2 R\&T} \\
    \cmidrule(lr){3-5}\cmidrule(lr){9-11}
    Model & Alg. & Small & Medium & Large & & & & Small & Medium & Large \\
    \midrule
    \multirow{3}{*}{GPT-5.4} & \tracer{} & \oodshare{10.8}{$-59.0$} & \oodshare{11.8}{$-52.5$} & \oodshare{12.7}{$-42.6$} & \oodshare{43.4}{$-35.4$} & \oodshare{2.3}{$-88.3$} & \oodshare{13.0}{$-32.5$} & \oodshare{33.0}{N/A} & \oodshare{40.4}{N/A} & \oodshare{36.7}{N/A} \\
     & \tracers{} & \oodshare{10.8}{$-47.1$} & \oodshare{11.8}{$-44.2$} & \oodshare{12.7}{$-40.5$} & \oodshare{43.4}{$-9.4$} & \oodshare{2.3}{$-57.4$} & \oodshare{13.0}{$-30.7$} & \oodshare{40.8}{N/A} & \oodshare{40.9}{N/A} & \oodshare{37.3}{N/A} \\
     & \trace{} & \oodshare{10.8}{$-55.9$} & \oodshare{11.8}{$-50.0$} & \oodshare{12.7}{$-34.4$} & \oodshare{43.4}{$-34.5$} & \oodshare{2.3}{$-91.4$} & \oodshare{13.0}{$-22.8$} & \oodshare{38.5}{N/A} & \oodshare{40.2}{N/A} & \oodshare{34.6}{N/A} \\
    \midrule
    \multirow{3}{*}{GPT-5.4-mini} & \tracer{} & \oodshare{10.8}{$-54.6$} & \oodshare{11.8}{$-46.7$} & \oodshare{12.7}{$-36.0$} & \oodshare{43.4}{$-34.5$} & \oodshare{2.3}{$-65.1$} & \oodshare{13.0}{$-36.2$} & \oodshare{35.6}{N/A} & \oodshare{43.6}{N/A} & \oodshare{39.2}{N/A} \\
     & \tracers{} & \oodshare{10.8}{$-42.3$} & \oodshare{11.8}{$-38.6$} & \oodshare{12.7}{$-32.7$} & \oodshare{43.4}{$-21.9$} & \oodshare{2.3}{$-32.5$} & \oodshare{13.0}{$-34.0$} & \oodshare{43.2}{N/A} & \oodshare{44.2}{N/A} & \oodshare{40.2}{N/A} \\
     & \trace{} & \oodshare{10.8}{$-51.2$} & \oodshare{11.8}{$-43.7$} & \oodshare{12.7}{$-29.5$} & \oodshare{43.4}{$-33.5$} & \oodshare{2.3}{$-64.3$} & \oodshare{13.0}{$-26.0$} & \oodshare{42.5}{N/A} & \oodshare{43.8}{N/A} & \oodshare{36.2}{N/A} \\
    \midrule
    \multirow{3}{*}{Grok-4-20-r} & \tracer{} & \oodshare{11.5}{$-59.3$} & \oodshare{12.0}{$-51.3$} & \oodshare{12.4}{$-39.5$} & \oodshare{43.4}{$-33.1$} & \oodshare{2.3}{$-65.1$} & \oodshare{12.9}{$-35.8$} & & & \\
     & \tracers{} & \oodshare{11.5}{$-49.6$} & \oodshare{12.0}{$-44.7$} & \oodshare{12.4}{$-36.5$} & \oodshare{43.4}{$-11.9$} & \oodshare{2.3}{$-31.8$} & \oodshare{12.9}{$-35.6$} & & & \\
     & \trace{} & \oodshare{11.5}{$-54.5$} & \oodshare{12.0}{$-47.1$} & \oodshare{12.4}{$-33.3$} & \oodshare{43.4}{$-32.5$} & \oodshare{2.3}{$-64.3$} & \oodshare{13.1}{$-24.8$} & & & \\
    \midrule
    \multirow{3}{*}{Mistral-Large-3} & \tracer{} & \oodshare{10.8}{$-60.2$} & \oodshare{11.8}{$-52.5$} & \oodshare{12.7}{$-43.1$} & \oodshare{43.4}{$-34.8$} & \oodshare{2.3}{$-28.6$} & \oodshare{13.0}{$-33.5$} & & & \\
     & \tracers{} & \oodshare{10.8}{$-47.5$} & \oodshare{11.8}{$-44.7$} & \oodshare{12.7}{$-39.7$} & \oodshare{43.4}{$-12.1$} & \oodshare{2.3}{$+3.9$} & \oodshare{13.0}{$-32.7$} & & & \\
     & \trace{} & \oodshare{10.8}{$-54.4$} & \oodshare{11.8}{$-49.1$} & \oodshare{12.7}{$-35.5$} & \oodshare{43.4}{$-34.1$} & \oodshare{2.3}{$-27.1$} & \oodshare{13.0}{$-22.3$} & & & \\
    \bottomrule
  \end{tabular}
  \par\smallskip
  {\footnotesize The large figure is the share of steps with no matching
   prefix node. Beneath it, in parentheses, is selection accuracy restricted
   to those steps minus the same cell's overall accuracy in
   Table~\ref{tab:main-tf}, in points. $k{=}10$ throughout.}
\end{table}

\begin{multicols}{2}
\subsection{Constructing the free-running environment}
\label{app:env}

Where a benchmark ships an executable environment we reuse it and rebuild only
the grader at turn granularity; where tools are synthetic we build the executor,
state store, and grader ourselves. Both tracks end at the same per-turn
reference (\S\ref{sec:setup-env}), so one grader serves the mixed environment,
routed per example at run time. We inherit API functions and $\tau^2$ databases
where they exist and author them where they do not. An executor given arguments
absent from the database returns a sentinel rather than raising, so logs still
identify the step at which a failure began.

\subsection{Banking accuracy by tool type}
\label{app:banktype}
In banking a wrong tool means a skipped prerequisite rather than a wasted call.
The gains concentrate on the three tool types where the correct call is not the
obvious one: \emph{gated-unlock} calls ($+0.098$, $0.148\!\to\!0.246$),
\emph{shell/grep} filesystem calls ($+0.069$), and \emph{knowledge-base}
retrievals ($+0.054$). Accuracy is given back on read, write, verification and
routing calls, for a net $-0.023$ overall.

\begin{table}[H]
  \centering
  \caption{\textbf{Banking accuracy by gold tool type} (GPT-5.4, $k{=}10$;
    $\Delta{=}\tracers{}-\llmonly{}$).}
  \label{tab:bank-tooltype}
  \small
  \setlength{\tabcolsep}{3pt}
  \begin{tabular}{@{}l r cc r@{}}
    \toprule
    Gold tool type & Steps & \tracers{} & \llmonly{} & $\Delta$ \\
    \midrule
    gated unlock\textsuperscript{$\ast$} & 228     & 0.246 & 0.148 & \up{$+0.098$} \\
    shell / grep                         & 1{,}143 & 0.553 & 0.484 & \up{$+0.069$} \\
    KB retrieval                         & 594     & 0.575 & 0.521 & \up{$+0.054$} \\
    write                                & 230     & 0.730 & 0.867 & \dn{$-0.136$} \\
    verification                         & 73      & 0.721 & 0.941 & \dn{$-0.219$} \\
    utility routing                      & 197     & 0.423 & 0.689 & \dn{$-0.266$} \\
    read                                 & 293     & 0.386 & 0.721 & \dn{$-0.336$} \\
    \midrule
    Overall (micro)                      & 2{,}758 & 0.525 & 0.548 & \dn{$-0.023$} \\
    \bottomrule
  \end{tabular}\\[3pt]
  {\footnotesize\textsuperscript{$\ast$}\,\emph{gated unlock} $=$ the
   \api{unlock:}$X$ call the agent must issue before a sensitive tool $X$.
   Rows sorted by $\Delta$; ``Steps'' counts gold decision steps of that type.}
\end{table}

\subsection{Full budget sweeps}
\label{app:fullgrid}
Table~\ref{tab:main-tf} in the body reports the $k{=}10$ operating point.
The complete sweeps over every budget $k$, for both selection accuracy and context-injection cost, are below.

\end{multicols}
\begin{table}[H]
  \centering
  \caption{\textbf{Teacher-Forcing tool-selection accuracy (\%): full budget
    sweep, Mixed Tau2 R\,\&\,T.}
    Parentheses give the across-fold SD. Best per column within each model in
    \textbf{bold}, second best \underline{underlined}. \mi{}: averaged over
    3 iterations.}
  \label{tab:main-tf-full}
  \scriptsize
  \setlength{\tabcolsep}{1.2pt}
  \begin{tabular}{@{}ll cccc cccc cccc@{}}
    \toprule
    & & \multicolumn{4}{c}{\shortstack{Small\\Mixed Tau2 R\,\&\,T}} & \multicolumn{4}{c}{\shortstack{Medium\\Mixed Tau2 R\,\&\,T}} & \multicolumn{4}{c}{\shortstack{Large\\Mixed Tau2 R\,\&\,T\mi}} \\
    \cmidrule(lr){3-6}\cmidrule(lr){7-10}\cmidrule(lr){11-14}
    Model & Alg. & $5$ & $10$ & $20$ & $30$ & $5$ & $10$ & $20$ & $30$ & $5$ & $10$ & $20$ & $30$ \\
    \midrule
    \multirow{5}{*}{GPT-5.4} & \llmonly{} & \runnerup{63.8}(3.1) & 64.1(3.3) & 64.0(3.5) & 64.1(2.9) & 60.2(3.5) & 60.1(3.4) & 60.1(3.5) & 60.0(3.5) & 57.2(1.2) & 57.3(1.1) & 57.3(1.2) & 57.5(1.4) \\
     & \tracer{} & \best{69.4}(1.5) & \best{69.3}(1.7) & \best{69.7}(2.5) & \best{69.4}(2.3) & \best{66.1}(2.9) & \best{63.7}(2.6) & \best{63.5}(3.3) & \best{63.1}(3.1) & \best{61.9}(1.6) & \best{59.0}(1.7) & \best{58.7}(1.8) & \best{59.0}(1.9) \\
     & \tracers{} & 62.4(1.5) & 62.9(2.3) & 63.0(2.7) & 62.7(2.5) & \runnerup{61.9}(3.3) & 60.4(3.4) & \runnerup{60.1}(3.4) & 59.7(3.7) & \runnerup{59.9}(1.7) & \runnerup{58.0}(1.8) & \runnerup{57.9}(1.5) & \runnerup{58.0}(2.1) \\
     & \trace{} & 62.2(1.3) & \runnerup{66.3}(1.8) & \runnerup{69.5}(1.5) & \runnerup{69.1}(2.5) & 60.3(2.4) & \runnerup{61.9}(2.8) & 59.5(3.7) & \runnerup{60.6}(3.2) & 50.5(1.1) & 51.0(2.2) & 53.6(2.2) & 55.5(2.1) \\
     & \vecsearch{} & 26.5(2.3) & 38.6(2.2) & 53.6(3.4) & 59.6(3.4) & 26.1(1.1) & 34.0(1.7) & 45.9(3.3) & 52.0(2.5) & 27.7(1.5) & 34.4(1.5) & 43.1(1.5) & 50.0(2.1) \\
    \midrule
    \multirow{5}{*}{GPT-5.4-mini} & \llmonly{} & 57.6(3.1) & 57.2(3.2) & 57.1(2.7) & 57.2(3.0) & 55.8(3.1) & 55.3(3.4) & 55.1(3.5) & 55.2(3.4) & 54.5(1.7) & \runnerup{54.7}(1.7) & \runnerup{54.7}(1.7) & \runnerup{54.6}(1.8) \\
     & \tracer{} & \best{67.5}(1.4) & \best{66.3}(2.3) & \best{65.4}(2.6) & \best{64.0}(2.6) & \best{62.6}(2.7) & \best{60.4}(2.6) & \best{59.6}(3.0) & \best{58.8}(3.0) & \best{58.5}(1.8) & \best{55.6}(1.8) & \best{55.0}(1.8) & \best{54.9}(2.3) \\
     & \tracers{} & 59.6(2.0) & 59.9(2.9) & 58.9(3.9) & 57.0(3.4) & \runnerup{58.8}(3.0) & 56.6(3.0) & \runnerup{56.3}(3.4) & 55.5(3.5) & \runnerup{56.5}(1.5) & 54.4(1.9) & 53.8(1.9) & 53.5(2.2) \\
     & \trace{} & \runnerup{60.5}(2.3) & \runnerup{63.4}(1.4) & \runnerup{64.9}(1.7) & \runnerup{63.7}(2.7) & 56.9(2.7) & \runnerup{57.4}(3.2) & 55.7(3.3) & \runnerup{56.4}(3.2) & 48.0(3.4) & 49.7(3.3) & 51.2(2.4) & 52.7(2.6) \\
     & \vecsearch{} & 25.3(2.7) & 36.0(2.6) & 47.6(3.7) & 52.4(3.8) & 26.4(1.4) & 32.9(2.5) & 41.8(3.5) & 45.5(3.3) & 28.4(1.8) & 33.5(1.7) & 40.9(2.3) & 45.5(2.5) \\
    \midrule
    \multirow{5}{*}{Grok-4-20-r} & \llmonly{} & 62.6(3.6) & 62.4(3.5) & 62.7(3.4) & 62.9(3.6) & 60.5(4.1) & 60.6(4.0) & 60.5(4.0) & 60.9(3.7) & \best{59.8}(1.5) & \best{60.2}(1.2) & \best{59.5}(1.4) & \best{60.4}(1.3) \\
     & \tracer{} & \best{70.4}(2.0) & \best{70.8}(1.9) & \best{70.7}(2.6) & \best{69.3}(2.5) & \best{66.2}(3.4) & \best{65.1}(2.8) & \best{64.8}(3.0) & \best{64.4}(3.9) & \runnerup{59.5}(3.8) & \runnerup{58.3}(3.8) & \runnerup{58.5}(3.7) & \runnerup{58.6}(4.3) \\
     & \tracers{} & \runnerup{65.7}(2.4) & \runnerup{66.1}(2.4) & 66.4(3.1) & 63.7(3.2) & \runnerup{63.9}(3.7) & \runnerup{62.8}(3.0) & \runnerup{62.5}(3.7) & 62.3(4.1) & 58.3(4.0) & 57.7(3.6) & 58.1(3.4) & 57.5(4.0) \\
     & \trace{} & 62.6(1.5) & 65.5(1.0) & \runnerup{68.3}(1.8) & \runnerup{67.3}(2.2) & 59.4(2.5) & 61.3(2.9) & 62.3(3.2) & \runnerup{62.5}(3.7) & 48.4(2.7) & 52.3(3.2) & 54.9(2.9) & 56.8(3.5) \\
     & \vecsearch{} & 25.5(1.1) & 35.8(1.3) & 48.0(2.5) & 53.7(2.6) & 26.6(0.9) & 34.3(1.4) & 45.3(3.0) & 50.3(3.3) & 27.4(2.6) & 32.8(2.3) & 42.2(2.5) & 48.1(3.1) \\
    \midrule
    \multirow{5}{*}{Mistral-Large-3} & \llmonly{} & \runnerup{66.1}(2.2) & 65.6(2.2) & 66.0(2.2) & 65.8(2.2) & 62.6(3.0) & 62.8(2.6) & \runnerup{62.4}(2.9) & \runnerup{62.6}(3.0) & 61.2(2.0) & \runnerup{61.2}(2.0) & \runnerup{60.8}(2.0) & \runnerup{60.9}(2.0) \\
     & \tracer{} & \best{71.6}(1.4) & \best{71.5}(1.7) & \best{70.7}(2.3) & \runnerup{70.0}(2.2) & \best{68.3}(2.5) & \best{65.5}(2.3) & \best{64.4}(2.9) & \best{64.0}(3.5) & \best{64.5}(2.2) & \best{62.2}(2.4) & \best{61.4}(2.7) & \best{61.1}(3.2) \\
     & \tracers{} & 65.5(1.9) & 67.0(2.0) & 65.8(3.4) & 65.2(2.8) & \runnerup{64.4}(3.3) & \runnerup{63.5}(3.3) & 61.9(3.4) & 62.2(4.0) & \runnerup{62.3}(2.1) & 60.5(2.4) & 59.9(2.5) & 60.0(3.0) \\
     & \trace{} & 64.2(1.4) & \runnerup{67.2}(1.7) & \runnerup{70.5}(2.4) & \best{70.0}(2.7) & 60.2(2.6) & 61.9(2.8) & 61.6(4.0) & 62.4(3.5) & 53.6(1.1) & 54.6(2.0) & 58.1(2.3) & 59.7(2.5) \\
     & \vecsearch{} & 26.5(2.3) & 39.2(2.4) & 54.5(3.4) & 61.2(3.4) & 26.6(1.3) & 34.1(1.7) & 47.0(3.1) & 52.3(2.9) & 27.3(1.7) & 32.5(1.7) & 42.9(2.5) & 50.6(2.3) \\
  \bottomrule
  \end{tabular}
\end{table}

\begin{table}[H]
  \centering
  \caption{\textbf{Teacher-Forcing tool-selection accuracy (\%): full budget
    sweep, Tau3~Banking, API-Bank and BFCL v3.}
    Parentheses give the across-fold SD. Best per column within each model in
    \textbf{bold}, second best \underline{underlined}. \mi{}: averaged over
    3 iterations. Tau3~Banking runs to $k{=}10$, API-Bank to $k{=}5$.}
  \label{tab:main-tf-full-b}
  \scriptsize
  \setlength{\tabcolsep}{1.2pt}
  \begin{tabular}{@{}ll ccc cc cccc@{}}
    \toprule
    & & \multicolumn{3}{c}{\shortstack{Tau3\\Banking\mi}} & \multicolumn{2}{c}{\shortstack{API-\\Bank}} & \multicolumn{4}{c}{BFCL v3} \\
    \cmidrule(lr){3-5}\cmidrule(lr){6-7}\cmidrule(lr){8-11}
    Model & Alg. & $3$ & $5$ & $10$ & $3$ & $5$ & $5$ & $10$ & $20$ & $30$ \\
    \midrule
    \multirow{5}{*}{GPT-5.4} & \llmonly{} & \best{54.7}(1.3) & \best{55.0}(2.4) & \best{54.8}(1.6) & 81.6(9.1) & 82.4(6.8) & 61.6(2.8) & 61.1(2.8) & 61.4(2.9) & 61.9(3.2) \\
     & \tracer{} & 44.3(3.5) & 45.7(4.2) & 46.8(4.2) & 90.6(5.7) & 88.3(7.5) & \best{68.4}(3.7) & \best{68.1}(3.6) & \best{68.5}(2.9) & \best{69.6}(3.0) \\
     & \tracers{} & \runnerup{50.4}(3.0) & \runnerup{51.6}(3.2) & \runnerup{52.6}(3.4) & \runnerup{90.7}(6.2) & \runnerup{90.7}(6.2) & \runnerup{65.4}(3.5) & \runnerup{65.6}(3.8) & \runnerup{66.4}(3.3) & \runnerup{67.1}(3.2) \\
     & \trace{} & 41.9(4.2) & 43.6(4.4) & 45.9(4.4) & \best{93.0}(3.7) & \best{91.4}(4.5) & 63.1(2.3) & 64.2(3.4) & 65.8(2.6) & 66.2(2.5) \\
     & \vecsearch{} & 7.4(2.1) & 9.5(3.0) & 12.5(2.7) & 68.5(8.4) & 72.7(5.1) & 56.4(2.2) & 59.0(2.9) & 60.2(2.9) & 62.3(3.5) \\
    \midrule
    \multirow{5}{*}{GPT-5.4-mini} & \llmonly{} & \runnerup{49.2}(4.9) & \runnerup{49.3}(3.6) & \runnerup{49.4}(3.8) & 80.1(3.2) & 81.1(10.9) & 62.3(2.6) & 62.9(1.7) & 62.5(1.9) & 61.9(2.7) \\
     & \tracer{} & 44.3(3.8) & 44.6(4.0) & 45.2(3.7) & \runnerup{98.4}(2.0) & \runnerup{98.4}(2.0) & \best{70.2}(2.8) & \best{70.0}(2.9) & \best{71.4}(3.2) & \best{71.2}(2.6) \\
     & \tracers{} & \best{52.8}(3.6) & \best{53.4}(4.7) & \best{52.4}(4.3) & \best{99.2}(1.6) & \best{99.2}(1.6) & \runnerup{66.5}(2.2) & \runnerup{67.1}(1.5) & \runnerup{68.4}(2.3) & 68.8(2.2) \\
     & \trace{} & 41.2(4.0) & 42.8(4.0) & 44.3(4.2) & 96.9(1.6) & 97.6(1.9) & 63.7(2.9) & 66.5(4.0) & 68.0(3.8) & \runnerup{68.9}(2.9) \\
     & \vecsearch{} & 6.8(2.0) & 9.0(2.8) & 11.1(2.6) & 70.9(5.7) & 76.2(4.7) & 56.0(2.6) & 60.1(3.7) & 62.6(3.6) & 64.9(3.8) \\
    \midrule
    \multirow{5}{*}{Grok-4-20-r} & \llmonly{} & \best{58.3}(2.3) & \best{58.6}(2.5) & \best{58.7}(2.6) & 86.3(6.8) & 86.4(3.4) & 60.5(2.9) & 59.8(2.5) & 60.4(3.0) & 60.0(2.5) \\
     & \tracer{} & 42.3(4.1) & 42.4(4.3) & 42.7(4.1) & \runnerup{98.4}(2.0) & \runnerup{98.4}(2.0) & \best{71.8}(3.1) & \best{72.4}(3.9) & \best{73.4}(2.9) & \best{73.0}(2.4) \\
     & \tracers{} & \runnerup{55.6}(4.3) & \runnerup{56.5}(4.6) & \runnerup{55.7}(4.2) & \best{99.2}(1.6) & \best{98.5}(1.9) & \runnerup{70.4}(4.3) & \runnerup{71.3}(4.5) & \runnerup{71.9}(3.4) & \runnerup{71.0}(4.6) \\
     & \trace{} & 39.3(4.2) & 40.4(4.1) & 42.1(3.7) & 97.6(1.9) & 97.6(1.9) & 66.8(4.1) & 65.3(4.2) & 65.4(3.2) & 65.8(2.5) \\
     & \vecsearch{} & 7.3(1.6) & 8.9(1.8) & 10.9(2.6) & 71.0(6.2) & 76.4(1.7) & 60.3(3.4) & 59.1(3.1) & 60.6(3.2) & 61.4(2.8) \\
    \midrule
    \multirow{5}{*}{Mistral-Large-3} & \llmonly{} & \runnerup{47.8}(2.4) & \runnerup{48.0}(2.3) & \runnerup{47.6}(2.2) & 70.1(7.4) & 67.2(7.5) & 62.5(3.5) & 62.6(3.8) & 62.6(3.7) & 63.0(3.8) \\
     & \tracer{} & 43.2(3.8) & 43.4(3.9) & 44.3(3.9) & \runnerup{95.3}(2.9) & \runnerup{95.3}(2.9) & \best{70.1}(3.9) & \best{68.6}(3.4) & \best{69.8}(4.9) & \best{70.5}(3.6) \\
     & \tracers{} & \best{50.9}(3.9) & \best{51.0}(4.0) & \best{51.6}(4.3) & \best{96.1}(3.5) & \best{96.1}(3.5) & \runnerup{67.5}(4.8) & \runnerup{68.3}(5.0) & \runnerup{69.7}(5.4) & \runnerup{69.8}(5.5) \\
     & \trace{} & 40.5(4.2) & 41.6(4.6) & 43.2(4.1) & 94.6(3.8) & 93.8(3.8) & 64.1(2.4) & 62.8(3.5) & 64.1(3.4) & 64.2(2.5) \\
     & \vecsearch{} & 7.1(1.8) & 8.7(2.3) & 10.2(2.7) & 68.0(10.2) & 69.6(10.1) & 56.9(1.9) & 57.3(2.9) & 59.9(2.6) & 60.7(2.6) \\
  \bottomrule
  \end{tabular}
\end{table}

\begin{table}[H]
  \centering
  \caption{\textbf{Context-injection cost (CIC), Teacher-Forcing: full budget
    sweep, Mixed Tau2 R\,\&\,T.}
    Reduction against \llmonly{} in the same column; in parentheses, mean
    prompt tokens per decision step in thousands. CIC is independent of the
    serving model, so one row per algorithm covers all four.}
  \label{tab:cost-full}
  \scriptsize
  \setlength{\tabcolsep}{2pt}
  \begin{tabular}{@{}l cccc cccc cccc@{}}
    \toprule
    & \multicolumn{4}{c}{\shortstack{Small\\Mixed Tau2 R\,\&\,T}} & \multicolumn{4}{c}{\shortstack{Medium\\Mixed Tau2 R\,\&\,T}} & \multicolumn{4}{c}{\shortstack{Large\\Mixed Tau2 R\,\&\,T\mi}} \\
    \cmidrule(lr){2-5}\cmidrule(lr){6-9}\cmidrule(lr){10-13}
    Alg. & $5$ & $10$ & $20$ & $30$ & $5$ & $10$ & $20$ & $30$ & $5$ & $10$ & $20$ & $30$ \\
    \midrule
    \llmonly{} & \ciccell{7.54}{0} & \ciccell{7.54}{0} & \ciccell{7.54}{0} & \ciccell{7.54}{0} & \ciccell{10.97}{0} & \ciccell{10.97}{0} & \ciccell{10.97}{0} & \ciccell{10.97}{0} & \ciccell{14.17}{0} & \ciccell{14.17}{0} & \ciccell{14.17}{0} & \ciccell{14.17}{0} \\
    \textbf{\tracer{}} & \ciccellb{3.36}{55} & \ciccellb{3.64}{52} & \ciccellb{4.06}{46} & \ciccellb{4.46}{41} & \ciccellb{3.09}{72} & \ciccellb{3.43}{69} & \ciccellb{3.98}{64} & \ciccellb{4.41}{60} & \ciccellb{2.81}{80} & \ciccellb{3.22}{77} & \ciccellb{3.89}{73} & \ciccellb{4.53}{68} \\
    \tracers{} & \ciccell{3.58}{52} & \ciccell{3.85}{49} & \ciccell{4.26}{44} & \ciccell{4.64}{38} & \ciccell{3.34}{70} & \ciccell{3.66}{67} & \ciccell{4.20}{62} & \ciccell{4.61}{58} & \ciccell{3.07}{78} & \ciccell{3.46}{76} & \ciccell{4.12}{71} & \ciccell{4.75}{66} \\
    \trace{} & \ciccell{3.39}{55} & \ciccell{3.67}{51} & \ciccell{4.07}{46} & \ciccell{4.46}{41} & \ciccell{3.12}{72} & \ciccell{3.50}{68} & \ciccell{4.05}{63} & \ciccell{4.43}{60} & \ciccell{2.79}{80} & \ciccell{3.32}{77} & \ciccell{3.99}{72} & \ciccell{4.68}{67} \\
    \vecsearch{} & \ciccell{3.54}{53} & \ciccell{4.09}{46} & \ciccell{5.10}{32} & \ciccell{6.11}{19} & \ciccell{3.21}{71} & \ciccell{3.74}{66} & \ciccell{4.73}{57} & \ciccell{5.68}{48} & \ciccell{2.91}{79} & \ciccell{3.44}{76} & \ciccell{4.46}{68} & \ciccell{5.46}{61} \\
    \bottomrule
  \end{tabular}
\end{table}

\begin{table}[H]
  \centering
  \caption{\textbf{Context-injection cost (CIC), Teacher-Forcing: full budget
    sweep, Tau3~Banking, API-Bank and BFCL v3.}
    Reduction against \llmonly{} in the same column; in parentheses, mean
    prompt tokens per decision step in thousands.}
  \label{tab:cost-full-b}
  \scriptsize
  \setlength{\tabcolsep}{2pt}
  \begin{tabular}{@{}l ccc cc cccc@{}}
    \toprule
    & \multicolumn{3}{c}{\shortstack{Tau3\\Banking\mi}} & \multicolumn{2}{c}{\shortstack{API-\\Bank}} & \multicolumn{4}{c}{BFCL v3} \\
    \cmidrule(lr){2-4}\cmidrule(lr){5-6}\cmidrule(lr){7-10}
    Alg. & $3$ & $5$ & $10$ & $3$ & $5$ & $5$ & $10$ & $20$ & $30$ \\
    \midrule
    \llmonly{} & \ciccell{40.57}{0} & \ciccell{40.57}{0} & \ciccell{40.57}{0} & \ciccell{9.37}{0} & \ciccell{9.37}{0} & \ciccell{12.46}{0} & \ciccell{12.46}{0} & \ciccell{12.46}{0} & \ciccell{12.46}{0} \\
    \textbf{\tracer{}} & \ciccell{27.92}{31} & \ciccell{28.10}{31} & \ciccell{28.50}{30} & \ciccellb{1.14}{88} & \ciccellb{1.23}{87} & \ciccellb{1.25}{90} & \ciccellb{1.63}{87} & \ciccellb{2.32}{81} & \ciccellb{3.02}{76} \\
    \tracers{} & \ciccellb{28.33}{30} & \ciccellb{28.48}{30} & \ciccellb{28.86}{29} & \ciccell{1.34}{86} & \ciccell{1.42}{85} & \ciccell{1.59}{87} & \ciccell{1.95}{84} & \ciccell{2.62}{79} & \ciccell{3.31}{73} \\
    \trace{} & \ciccell{27.92}{31} & \ciccell{28.10}{31} & \ciccell{28.47}{30} & \ciccell{1.15}{88} & \ciccell{1.25}{87} & \ciccell{1.28}{90} & \ciccell{1.66}{87} & \ciccell{2.37}{81} & \ciccell{3.06}{75} \\
    \vecsearch{} & \ciccell{28.19}{31} & \ciccell{28.50}{30} & \ciccell{29.25}{28} & \ciccell{1.37}{85} & \ciccell{1.65}{82} & \ciccell{1.37}{89} & \ciccell{1.89}{85} & \ciccell{2.86}{77} & \ciccell{3.79}{70} \\
    \bottomrule
  \end{tabular}
\end{table}


\begin{table}[H]
  \centering
  \caption{\textbf{Free-running task success (\%).}}
  \label{tab:main-fr}
  \scriptsize
  \setlength{\tabcolsep}{1.0pt}
  \begin{tabular}{@{}ll cccc cccc cccc@{}}
    \toprule
    & & \multicolumn{4}{c}{\shortstack{Mixed Tau2\\Small}} & \multicolumn{4}{c}{\shortstack{Mixed Tau2\\Medium}} & \multicolumn{4}{c}{\shortstack{Mixed Tau2\\Large}} \\
    \cmidrule(lr){3-6}\cmidrule(lr){7-10}\cmidrule(lr){11-14}
    Model & Alg. & $5$ & $10$ & $20$ & $30$ & $5$ & $10$ & $20$ & $30$ & $5$ & $10$ & $20$ & $30$ \\
    \midrule
    \multirow{5}{*}{GPT-5.4} & \llmonly{} & \best{47.6}(5.1) & \best{47.2}(4.5) & 46.0(5.1) & \runnerup{48.3}(4.0) & \best{47.4}(3.2) & \best{47.5}(4.5) & \best{48.6}(4.5) & 48.3(5.2) & \best{58.2}(2.0) & \runnerup{57.9}(2.3) & \runnerup{57.9}(1.5) & 58.2(2.2) \\
     & \tracer{} & 43.0(3.5) & 45.0(5.5) & \best{48.3}(4.5) & \best{48.5}(5.5) & 43.0(3.8) & 46.2(3.3) & \runnerup{48.2}(3.8) & \runnerup{49.5}(2.3) & 55.3(2.2) & 56.2(1.8) & 57.6(2.8) & 58.3(2.3) \\
     & \tracers{} & \runnerup{44.7}(4.4) & \runnerup{46.1}(4.8) & \runnerup{46.7}(4.5) & 46.7(5.7) & 44.2(3.4) & \runnerup{46.7}(3.2) & 47.7(3.7) & 49.0(2.8) & 55.6(1.4) & 56.6(2.0) & 57.3(2.7) & \runnerup{58.5}(2.3) \\
     & \trace{} & 39.8(5.1) & 43.0(5.6) & 45.9(3.4) & 46.5(3.8) & \runnerup{45.2}(4.0) & 44.0(3.4) & 47.7(4.1) & \best{49.7}(3.9) & \runnerup{56.2}(1.9) & \best{58.1}(2.7) & \best{58.5}(2.8) & \best{58.6}(2.8) \\
     & \vecsearch{} & 28.1(3.1) & 35.1(5.4) & 37.1(4.1) & 41.9(5.6) & 25.3(2.5) & 32.1(3.7) & 37.6(6.3) & 40.7(6.0) & 41.5(0.6) & 45.6(0.8) & 50.6(1.6) & 52.6(1.1) \\
    \midrule
    \multirow{5}{*}{GPT-5.4-mini} & \llmonly{} & 38.8(2.6) & 39.3(3.0) & 39.5(3.1) & 40.0(2.6) & 38.7(5.0) & 38.7(3.9) & 37.8(4.4) & 38.9(4.0) & 39.4(1.6) & 39.1(2.4) & 39.2(2.2) & 39.3(2.3) \\
     & \tracer{} & \runnerup{39.1}(3.4) & \runnerup{41.5}(5.1) & \runnerup{40.9}(3.6) & 40.2(3.9) & \runnerup{41.5}(3.3) & \runnerup{40.1}(3.4) & \runnerup{41.7}(4.0) & \best{43.3}(3.3) & 53.1(2.2) & \runnerup{52.9}(1.9) & \best{54.4}(2.7) & 53.7(1.3) \\
     & \tracers{} & \best{40.2}(4.3) & \best{41.8}(4.4) & 40.9(4.9) & \runnerup{41.9}(3.2) & 40.9(3.2) & \best{40.5}(4.1) & \best{42.0}(3.8) & 42.7(3.4) & \best{53.7}(2.7) & 52.7(1.7) & \runnerup{53.7}(2.1) & \best{54.3}(2.6) \\
     & \trace{} & 34.5(4.5) & 39.1(3.9) & \best{40.9}(3.4) & \best{42.0}(3.6) & \best{42.4}(2.7) & 38.4(3.9) & 40.7(3.4) & \runnerup{42.9}(3.9) & \runnerup{53.4}(2.3) & \best{53.9}(1.9) & 53.5(1.7) & \runnerup{54.2}(1.7) \\
     & \vecsearch{} & 25.9(2.0) & 33.7(3.5) & 35.3(3.9) & 36.0(4.3) & 25.0(2.7) & 28.6(3.2) & 32.6(5.3) & 35.8(3.9) & 40.4(0.8) & 44.3(1.1) & 47.2(1.7) & 48.8(1.2) \\
    \bottomrule
  \end{tabular}
  \par\smallskip
  {\footnotesize Parentheses give the across-fold SD over the five
   evaluation folds.}
\end{table}
\begin{table}[H]
  \centering
  \caption{\textbf{Context-injection cost (CIC), Free-Running: full grid.}
    Reduction against \llmonly{} in the same model-and-registry row block; in
    parentheses, mean prompt tokens per decision step in thousands. Unlike
    teacher forcing, free-running CIC differs by serving model.}
  \label{tab:cic-fr}
  \scriptsize
  \setlength{\tabcolsep}{2pt}
  \begin{tabular}{@{}ll cccc cccc cccc@{}}
    \toprule
    & & \multicolumn{4}{c}{\shortstack{Small\\Mixed Tau2 R\,\&\,T}} & \multicolumn{4}{c}{\shortstack{Medium\\Mixed Tau2 R\,\&\,T}} & \multicolumn{4}{c}{\shortstack{Large\\Mixed Tau2 R\,\&\,T}} \\
    \cmidrule(lr){3-6}\cmidrule(lr){7-10}\cmidrule(lr){11-14}
    Model & Alg. & $5$ & $10$ & $20$ & $30$ & $5$ & $10$ & $20$ & $30$ & $5$ & $10$ & $20$ & $30$ \\
    \midrule
    \multirow{5}{*}{GPT-5.4} & \llmonly{} & \ciccell{6.91}{0} & \ciccell{6.48}{0} & \ciccell{6.88}{0} & \ciccell{6.62}{0} & \ciccell{11.45}{0} & \ciccell{11.43}{0} & \ciccell{11.42}{0} & \ciccell{11.42}{0} & \ciccell{11.99}{0} & \ciccell{11.97}{0} & \ciccell{11.98}{0} & \ciccell{11.98}{0} \\
     & \tracer{} & \ciccell{3.31}{52} & \ciccell{3.91}{40} & \ciccell{4.28}{38} & \ciccell{5.39}{19} & \ciccell{3.61}{68} & \ciccell{4.07}{64} & \ciccell{4.88}{57} & \ciccell{5.69}{50} & \ciccell{2.75}{77} & \ciccell{3.12}{74} & \ciccell{3.87}{68} & \ciccell{4.63}{61} \\
     & \tracers{} & \ciccell{3.30}{52} & \ciccell{3.94}{39} & \ciccell{4.37}{36} & \ciccell{5.36}{19} & \ciccell{3.96}{65} & \ciccell{4.32}{62} & \ciccell{5.10}{55} & \ciccell{5.87}{49} & \ciccell{2.98}{75} & \ciccell{3.35}{72} & \ciccell{4.09}{66} & \ciccell{4.81}{60} \\
     & \trace{} & \ciccell{3.48}{50} & \ciccell{4.15}{36} & \ciccell{4.93}{28} & \ciccell{5.29}{20} & \ciccell{3.61}{68} & \ciccell{4.04}{65} & \ciccell{4.86}{57} & \ciccell{5.67}{50} & \ciccell{2.70}{77} & \ciccell{3.07}{74} & \ciccell{3.81}{68} & \ciccell{4.56}{62} \\
     & \vecsearch{} & \ciccell{3.00}{57} & \ciccell{3.40}{48} & \ciccell{4.15}{40} & \ciccell{5.00}{24} & \ciccell{3.57}{69} & \ciccell{4.08}{64} & \ciccell{5.17}{55} & \ciccell{6.12}{46} & \ciccell{2.71}{77} & \ciccell{3.15}{74} & \ciccell{4.03}{66} & \ciccell{4.88}{59} \\
    \midrule
    \multirow{5}{*}{GPT-5.4-mini} & \llmonly{} & \ciccell{6.23}{0} & \ciccell{6.42}{0} & \ciccell{6.59}{0} & \ciccell{6.11}{0} & \ciccell{11.44}{0} & \ciccell{11.49}{0} & \ciccell{11.52}{0} & \ciccell{11.49}{0} & \ciccell{11.89}{0} & \ciccell{11.88}{0} & \ciccell{11.91}{0} & \ciccell{11.89}{0} \\
     & \tracer{} & \ciccell{3.31}{47} & \ciccell{3.69}{42} & \ciccell{4.43}{33} & \ciccell{5.21}{15} & \ciccell{3.56}{69} & \ciccell{4.00}{65} & \ciccell{4.89}{58} & \ciccell{5.70}{50} & \ciccell{2.67}{78} & \ciccell{3.08}{74} & \ciccell{3.82}{68} & \ciccell{4.58}{61} \\
     & \tracers{} & \ciccell{3.27}{48} & \ciccell{3.85}{40} & \ciccell{4.30}{35} & \ciccell{5.25}{14} & \ciccell{3.89}{66} & \ciccell{4.26}{63} & \ciccell{5.08}{56} & \ciccell{5.95}{48} & \ciccell{2.95}{75} & \ciccell{3.31}{72} & \ciccell{4.07}{66} & \ciccell{4.77}{60} \\
     & \trace{} & \ciccell{3.49}{44} & \ciccell{3.97}{38} & \ciccell{4.99}{24} & \ciccell{5.76}{6} & \ciccell{3.56}{69} & \ciccell{4.02}{65} & \ciccell{4.86}{58} & \ciccell{5.65}{51} & \ciccell{2.65}{78} & \ciccell{3.03}{74} & \ciccell{3.82}{68} & \ciccell{4.57}{62} \\
     & \vecsearch{} & \ciccell{3.11}{50} & \ciccell{3.54}{45} & \ciccell{4.46}{32} & \ciccell{5.61}{8} & \ciccell{3.48}{70} & \ciccell{3.98}{65} & \ciccell{5.02}{56} & \ciccell{6.02}{48} & \ciccell{2.63}{78} & \ciccell{3.08}{74} & \ciccell{3.93}{67} & \ciccell{4.80}{60} \\
    \bottomrule
  \end{tabular}
\end{table}

\begin{multicols}{2}
\subsection{Use of AI assistants}
\label{app:aiuse}
We used an AI coding assistant for software development. The manuscript was
drafted by the authors; an AI assistant subsequently edited parts of that text,
and was used for \LaTeX{} formatting and layout throughout and to draft two
appendix sections, \S\ref{app:humanstudy} and \S\ref{app:compute}, from facts
supplied by the authors. Experimental design, results, and every reported number
are the authors' own and were verified by the authors. AI systems also appear in
this work as objects of study, namely the four evaluated models, the frozen
embedder, and the LLM judges of the curation battery (\S\ref{app:curation});
those are components of the method rather than writing assistance.
\end{multicols}

\end{document}